\documentclass[final,times]{elsarticle}

\usepackage{amssymb}
\usepackage{amsmath}
\usepackage{comment}

\usepackage{graphicx}               
\usepackage{amsmath}                
\usepackage[colorlinks=true, allcolors=blue]{hyperref} 
\usepackage[numbers]{natbib}     
\usepackage{subcaption}             
\usepackage{multirow}               
\usepackage{algorithm}              
\usepackage[english]{babel}         
\usepackage{pifont}
\usepackage{tikz}
\usepackage[noend]{algpseudocode}\newcommand{\vect}[1]{\mathbf{#1}} 
\newcommand{\mat}[1]{\mathbf{#1}} 
\newcommand{\set}[1]{\mathcal{#1}} 
\usepackage{booktabs}
\usepackage{tabularx}
\usepackage{siunitx}
\usepackage{makecell}

\newcolumntype{L}{>{\raggedright\arraybackslash}X}
\journal{Automation in Construction}
\date{Accepted for publication}
\def\checkmark{\tikz\fill[scale=0.4](0,.35) -- (.25,0) -- (1,.7) -- (.25,.15) -- cycle;}

\begin{document}

\begin{frontmatter}



\title{Deep Learning-based Scalable Image-to-3D Facade Parser for Generating Thermal 3D Building Models}


\author[chalmers_cse]{Yinan Yu}
\footnote{Corresponding author: Yinan Yu. Email: yinan@chalmers.se}
\author[chalmers_ace]{Alex Gonzalez-Caceres}
\author[asymptotic_ai]{Samuel Scheidegger}
\author[chalmers_ace]{Sanjay Somanath}
\author[chalmers_ace]{Alexander Hollberg}

\affiliation[chalmers_cse]{
    organization={Department of Computer Science and Engineering, Chalmers University of Technology}, 
    city={Gothenburg},
    postcode={412 96}, 
    country={Sweden}
}

\affiliation[chalmers_ace]{
    organization={Department of Architecture and Civil Engineering, Chalmers University of Technology}, 
    city={Gothenburg},
    postcode={412 96}, 
    country={Sweden}
}

\affiliation[asymptotic_ai]{
    organization={Asymptotic AI}, 
    city={Gothenburg},
    country={Sweden}
}

\begin{abstract}
Renovating existing buildings is essential for climate impact. Early-phase renovation planning requires simulations based on thermal 3D models at Level of Detail (LoD) 3, which include features like windows. However, scalable and accurate identification of such features remains a challenge. This paper presents the Scalable Image-to-3D Facade Parser (SI3FP), a pipeline that generates LoD3 thermal models by extracting geometries from images using both computer vision and deep learning. Unlike existing methods relying on segmentation and projection, SI3FP directly models geometric primitives in the orthographic image plane, providing a unified interface while reducing perspective distortions. SI3FP supports both sparse (e.g., Google Street View) and dense (e.g., hand-held camera) data sources. Tested on typical Swedish residential buildings, SI3FP achieved approximately 5\% error in window-to-wall ratio estimates, demonstrating sufficient accuracy for early-stage renovation analysis. The pipeline facilitates large-scale energy renovation planning and has broader applications in urban development and planning.
\end{abstract}

\begin{highlights}
\item Scalable pipeline to generate 3D thermal models at LoD3 using sparse and dense data.
\item Orthographic transformation corrects distortions for accurate scale measurement.
\item Uses orthographic images as a unified interface, improving pipeline usability.
\item Achieves $\sim$5\% error in Window-to-Wall Ratio (WWR) estimation for energy renovation.
\item Enables large-scale building modeling for renovation and urban planning.
\end{highlights}


\end{frontmatter}



\section{Introduction}

In the pursuit of mitigating climate change, the renovation of existing buildings plays a key role. Bringing buildings up to modern energy standards is one of the key strategies to reduce energy consumption and greenhouse gas emissions \citep{galimshina_strategies_2024}. This is especially true for European apartment buildings built in the 1970s and earlier, before energy performance regulations were introduced \citep{BALARAS2005515}. Energy simulation is most commonly used to evaluate potential renovation alternatives before implementation and to support decision-makers in picking the right one. As decision-makers are often building owners of large portfolios, including hundreds of buildings, there is a need for efficient and scalable solutions for evaluating this building stock. Central to any building energy simulation is a thermal 3D model of the building that includes its most important heat transfer elements, such as walls, roofs, and windows. Unfortunately, as many older buildings lack up-to-date floor plans or CAD files \citep{Bizjak2021}, energy modellers often turn to municipal or cadastral data, which rarely exceed Level of Detail (LoD2)- which represent buildings as simple block models without facade details like windows.
LoD3.0 models, on the other hand, capture roof details with higher accuracy while other features remain at a lower level of detail.
Although LoD3.0 models already contain computational information for various applications, the taxonomy and refined structures of windows are often ignored \citep{zhu_structure-aware_2021}. 
However, including windows in thermal 3D modeling is crucial for calculating solar gains and inaccuracies in estimating the buildings envelope can result in incorrect evaluations of retrofit measures. \cite{MaAnalysisOfw2wr} showed that halving the Window-to-Wall Ratio (WWR), from 0.28 to 0.14, cuts heating loads by 9.4 \%–13.3 \%, highlighting the strong linear sensitivity of energy demand to WWR.
The required level of detail that includes windows is defined as LoD3.1 according to~\cite{biljecki_improved_2016}.

Existing solutions fall into two categories: top-down and bottom-up approaches. 
Top-down approaches, such as the use of archetype libraries like IEE-TABULA~\citep{Loga2016}, model cities using average buildings but underestimate real-world variety~\citep{Mohammadiziazi2021}, making them unsuitable for informing building-level renovations~\citep{Osterbring2016}. 
Bottom-up approaches, such as the physical modeling method~\citep{Li2020}, rely on detailed thermodynamic modeling of individual buildings~\citep{AydinalpKoksal2008, Fonseca2015}. LiDAR scanning is a key technology here, offering high-precision 3D data. However, LiDAR has notable drawbacks: high equipment and processing costs, lack of semantic information, and reliance on labor-intensive manual annotation~\citep{li_less_2023}. These barriers limit its scalability for large-scale building portfolio analysis.

To address these challenges, researchers have explored cameras as primary 3D reconstruction sensors. Cameras are low-cost, flexible, and benefit from advances in computer vision and deep learning (see Section~\ref{sec:related_work}). They enable semantic analysis through object detection and instance segmentation. However, camera-based methods also face limitations, including perspective distortion, occlusions, and the need for dense image collections for reliable 3D reconstruction.

\paragraph{Research gap} Despite progress, to our knowledge, there is no end-to-end, fully automated pipeline that generates LoD3.1 building models from images with semantically accurate window and facade details. Top-down archetypes are too generalized for individual building renovations, and LiDAR-based approaches, while accurate, are costly and require extensive manual annotation and post-processing. Hence, there remains a critical need for a scalable, cost-effective method to generate LoD3.1 models with sufficient geometric and semantic detail for early-stage energy renovation analysis.

The aim of this paper is to develop a pipeline that generates thermal LoD3.1 models using scalable image-based data sources. Scalability considers not only equipment cost and data collection effort but also the complexity of data processing. Our goal is to support building portfolio owners in selecting optimal renovation solutions efficiently.

The overarching research question we address is:
\begin{center}
\emph{Which data sources and algorithms can be applied to generate thermal 3D models for building energy renovation with LoD3-level detail and sufficient accuracy?}
\end{center}

We propose a camera-based pipeline demonstrated on three residential buildings from the 1961–1975 period, a representative case for energy renovation in Europe. Our design minimizes the need for annotated training data and model fine-tuning across different built environments. 

\begin{figure}[ht]
  \centering
  \includegraphics[width=1\linewidth]{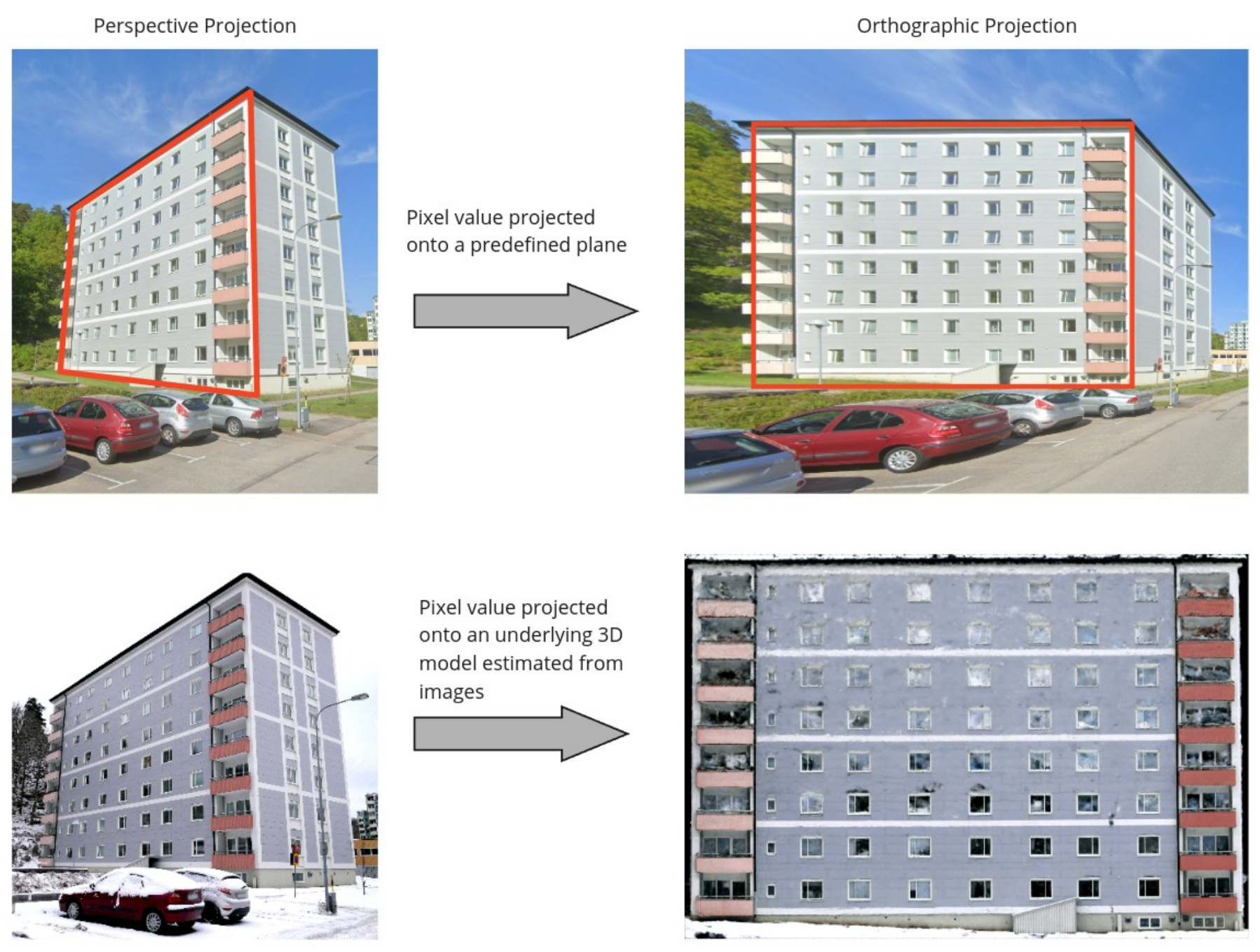}
  \caption{\label{fig:orthographic} Comparison of perspective images with proposed orthographic images for both sparse images (first row; from Google Street View) and dense images (second row; using a perspective camera).}
\end{figure}

Our main contributions are as follows:
\begin{enumerate}
  \item We propose a unified framework that accommodates both sparse and dense data collections, enabling flexible and scalable 3D facade modeling under varying data availability conditions.
  \item For sparse street-level imagery (e.g., Google Street View), we introduce an ensemble-based fusion method (Algorithm 6) that aggregates multiple partial orthographic views to improve robustness against occlusions, viewpoint variations, and localization noise.
  \item For dense image collections, we leverage Neural Radiance Fields (NeRF) not only for 3D reconstruction but specifically for generating detailed facade renderings, allowing direct computation of true orthographic images via surface-based parallel projection.
  \item Across both sparse and dense workflows, we apply orthographic transformations to standardize the data representation, simplify feature detection and geometry parameterization, and improve dimensional accuracy by mitigating perspective distortion (illustrated in Figure~\ref{fig:orthographic}). The orthographic transformation minimizes perspective distortion and simplifies the parameterization of facade features.
\end{enumerate}

Our pipeline effectively bridges the gap between scalability and accuracy in generating LoD3.1 building models required for thermal energy simulation, aiming to support decision-makers in evaluating renovation options and contribute to the broader goal of reducing energy consumption and mitigating climate change.

\section{Background and related works}

To select the most suitable sensors to enable scalable and efficient data collection, we first describe the relevant background on different sensors. We then review related works using these sensors for facade analysis and the generation of 3D models.

\subsection{Sensor types}
We categorize the sensor choices into two types: primary sensor and complementary information.
The primary sensor is the main system used for data collection, while the complementary information includes additional data sources that enhance the accuracy and detail of the 3D models.

\subsubsection{Primary sensor}

We differentiate between three commonly used sensor types: 1) LiDAR, 2) mono cameras, and 3) stereo cameras.

\paragraph{LiDAR (Light Detection and Ranging)}
LiDAR technology \citep{maksymova2018review, li2020lidar} stands out for its high precision in mapping the 3D structure of buildings, offering relatively quick data acquisition over large areas, which makes it a suitable choice for both small and large-scale projects. Further, it performs well under various lighting conditions.
However, it is challenging to extract semantic information from 3D data due to the inherently complexity of three-dimensional pattern recognition.
Moreover, the nature of 3D point clouds, often characterized by their sparsity, further complicates the task of semantic analysis. This is especially problematic when measuring transparent objects such as windows or dark objects.
In addition, the high cost of LiDAR systems may limit its accessibility for smaller projects.

\paragraph{(Mono) camera}

Using cameras for 3D reconstruction offers an accessible and cost-effective method for creating detailed 3D models of buildings, leveraging the widespread availability of consumer-grade devices. Cameras excel at capturing facade textures and colors, enabling semantic analysis such as object detection~\citep{zou2023object, zhao2019object} and instance segmentation~\citep{hafiz2020survey}. Flexible deployment options, including handheld and drone-mounted setups, further enhance their utility across diverse project scales.

\begin{itemize}
  \item {\bf Perspective images:} 
        3D reconstruction from video sequences is well-established in computer vision and photogrammetry~\citep{Hartley2004}. Recent advances in Artificial Intelligence (AI) and Deep Learning (DL) have significantly improved both visual quality and geometric accuracy. Dense video capture around a structure allows each frame to serve as a perspective image from a distinct camera pose.
  \item {\bf Panoramic images:} 
        Panoramic images \citep{gledhill2003panoramic, 10.1145/3365610.3365645}, captured using wide-angle lenses, multi-camera rigs, or rotating systems, collect rays from a broad field of view, enabling efficient coverage of large facades with fewer captures.
  \item {\bf Orthographic images:}
        Orthographic images \citep{szeliski2022computer, 10.1145/1463434.1463465} use parallel projection to represent scenes without perspective distortion, preserving true scale and geometry. They are particularly useful for accurate geometric measurements and facade element parameterization. However, standard cameras inherently capture perspective images, where light rays converge at a focal point, causing distortion with distance. As a result, true orthographic views are not captured directly and are typically generated through post-processing of reconstructed 3D models.
\end{itemize}

Despite their advantages, camera-based methods face challenges. High-quality capture depends heavily on lighting conditions, and post-processing is often required to produce accurate 3D models. Artifacts such as rolling shutter effects and motion blur further complicate reconstruction. Additionally, orthographic projections are sensitive to the quality of the underlying 3D model, making robust reconstruction critical when indirect computation is required.

\paragraph{Stereo camera}
These cameras emulate human binocular vision to capture 3D data, providing an accessible and cost-effective alternative to LiDAR~\citep{barnard1982computational, furukawa2015multi, HAMID20221663}. They are valued for their affordability, simplicity, and versatility, supporting both indoor and outdoor mapping with relatively straightforward data processing. However, stereo cameras offer lower accuracy than LiDAR, particularly over long distances, and their performance can degrade under poor lighting conditions or on reflective or smooth surfaces.

\paragraph{Summary}
LiDAR excels at capturing structural dimensions but struggles to differentiate scene elements, such as windows versus walls, without additional data layers, and its high cost can be prohibitive. Stereo cameras provide direct scale but require a more complex setup and can suffer from reduced accuracy, especially over long distances. In contrast, monocular cameras, combined with scale estimation techniques, offer a flexible and cost-effective solution for capturing detailed spatial data. For thermal 3D modeling, monocular cameras provide the necessary resolution, detail, and operational flexibility, making them a preferred choice for efficient, low-cost model generation.

\subsubsection{Complementary 3D information}\label{sec:camera2d_3d}

Densely collected monocular images can generate 3D point clouds via photogrammetry, but these lack absolute scale, which is crucial for 3D modeling. To address this, external scale information must be integrated. In this section, we introduce common sources for obtaining such scale data.

\paragraph{Ground Control Points (GCP)} 
GCPs are specific, accurately surveyed points on the Earth's surface used to georeference aerial or satellite imagery to real-world coordinates~\citep{villanueva2019optimization}. They are critical for ensuring positional accuracy in mapping projects and are typically established through GNSS or traditional surveying methods.
\paragraph{Local Coordinate Reference Points} 
Real-world spatial dimensions in the image plane can be estimated by measuring distances and angles between camera centers of registered images. These measurements enable a 3D similarity transformation to align the reconstructed model with a coordinate system, using known relative distances and angles as local reference points.

  \paragraph{Known 3D structures} This approach uses known 3D structures, such as flat surfaces (planes), within a model's 3D space. Examples include photographed objects like building walls or ground surfaces with known real-world dimensions. Accurate real-world measurements and spatial relationships can be extracted from images using this information. Although this method assumes local flatness, which may not always hold for curved or uneven surfaces, it provides a simplified and effective way to represent large-scale urban environments.

\paragraph{Camera position and pose}
When a localization sensor (e.g. a GNSS receiver) is available during the image capturing process, each image frame taken from the video sequence has an associated position in space, and the images can be anchored in a world coordinate system in 3D.
This anchoring allows for the alignment and integration of multiple images into a coherent and unified 3D model, where the scale is consistent with the real-world.
However, for data collection with low cost, an accurate localization sensor is often unavailable, making the scale estimation challenging.
In such cases, a manual localization process can be employed.

\subsection{Review of related work}\label{sec:related_work}

\begin{table*}[ht]
  \centering
  \resizebox{\linewidth}{!}{
\begin{tabular}{cp{20em}cccccccccc}
\hline
\textbf{Method} & \textbf{Pipeline} & \textbf{I} & \textbf{II} & \textbf{III} & \textbf{IV} & \textbf{V} & \textbf{VI} & \textbf{VII} & \textbf{VIII} & \textbf{IX} & \textbf{X} \\
\hline
\cite{fleet_learning_2014} & II $\rightarrow$ I $\rightarrow$ VIII (point cloud, mesh) + IV $\rightarrow$ IX (semantic segmentation) & $\checkmark$ & $\checkmark$ &  & $\checkmark$ &  &  &  & $\checkmark$ & $\checkmark$ &  \\
\hline
\cite{lotte_3d_2018} & II $\rightarrow$ VIII (point cloud, mesh) + IV $\rightarrow$ IX (semantic segmentation) &  & $\checkmark$ &  & $\checkmark$ &  &  &  & $\checkmark$ & $\checkmark$ &  \\
\hline
\cite{nishida_procedural_2018} & III + building silhouette $\rightarrow$ VIII (building mass grammar) + V $\rightarrow$ X (facade and window grammar) &  &  & $\checkmark$ &  & $\checkmark$ &  &  & $\checkmark$ &  & $\checkmark$ \\
\hline
\cite{bacharidis2018fusing} & Georeferenced stereoscopic images I $\rightarrow$ VIII + V $\rightarrow$ X & $\checkmark$ &  &  &  & $\checkmark$ &  &  & $\checkmark$ &  & $\checkmark$ \\
\hline
\cite{malihi_window_2018} & II  $\rightarrow$ VIII $\rightarrow$ IX &  & $\checkmark$ &  &  &  &  &  &$\checkmark$  & $\checkmark$ &  \\
\hline
\cite{bacharidis_3d_2020} & III $\rightarrow$ VIII (depth estimation) $\rightarrow$ IX (3D rendering, semantic segmentation) &  &  & $\checkmark$ &  &  &  &  & $\checkmark$ & $\checkmark$ &  \\
\hline
\cite{pantoja-rosero_generating_2022} & II $\rightarrow$ VIII (estimate planes $\rightarrow$ LoD2) + IV $\rightarrow$ IX (in images) + X (in 3D geometries) &  & $\checkmark$ &  & $\checkmark$ &  &  &  & $\checkmark$ & $\checkmark$ & $\checkmark$  \\
\hline
\cite{ward_estimating_2023} & II $\rightarrow$ VIII (mesh, RGB from images) + IV $\rightarrow$ IX  (in images) $\rightarrow$ X (in point clouds) &  & $\checkmark$ &  & $\checkmark$ &  &  &  & $\checkmark$ & $\checkmark$ & $\checkmark$ \\
\hline
\cite{salehitangrizi_3d_2024} & II $\rightarrow$ VIII $\rightarrow$ III $\rightarrow$ IX $\rightarrow$ X &  & $\checkmark$ & $\checkmark$ &  &  &  &  & $\checkmark$ & $\checkmark$ & $\checkmark$ \\
\hline
\cite{zhang_slod2win_2024} & VII + V $\rightarrow$ X (window detection and details for defining grammar) &  &  &  &  & $\checkmark$ &  & $\checkmark$ &  &  & $\checkmark$ \\
\hline 
\cite{PAL2024105157} & II + VII $\rightarrow$ VIII (SfM cloud/poses, NeRF model) + IV $\rightarrow$ \{V (Proj. Transf.) \textbar{} VI (NeRF Trav.)\} $\rightarrow$ IX (Semantic Segmentation) &  & $\checkmark$ &  & $\checkmark$ & $\checkmark$ & $\checkmark$ & $\checkmark$ & $\checkmark$ & $\checkmark$ &  \\
  \hline
\cite{tarkhan_facade_2024} & I + VII $\rightarrow$ V $\rightarrow$ X &$\checkmark$  & &&&$\checkmark$&& $\checkmark$&& & $\checkmark$\\\hline
SI3FP (ours) StreetView & I + VII (planes) $\rightarrow$ IV $\rightarrow$ VI $\rightarrow$ X (window detection) & $\checkmark$ &  &  & $\checkmark$ &  & $\checkmark$ &  $\checkmark$ & &  & $\checkmark$ \\
\hline
SI3FP (ours) Camera2D & II $\rightarrow$ VIII + IV $\rightarrow$ V $\rightarrow$ X (window detection) &  & $\checkmark$ &  &$\checkmark$  & $\checkmark$ &  &  & $\checkmark$ &   & $\checkmark$ \\
\hline
\end{tabular}}
\caption{Summary of related work (camera image as primary input for 3D modeling) and their respective processes and outcomes.}\label{tab:summary}
\end{table*}

There exist a few review papers on the topic of feature extraction using cameras and LiDAR sensors \citep{parente_integration_2023, musialski_survey_2013, klimkowska_detailed_2022, li_street_2022}. In this section, we briefly give an overview of this field.
\subsubsection{Facade analysis in the camera image plane}

Computer vision and image analysis are fundamental building blocks for facade parsing, as images contain rich semantic information through their pixel values. While analysis conducted solely in the image plane is insufficient for thermal 3D modeling, it remains essential in most facade parsing pipelines. In this section, we introduce articles that focus only on the image plane for facade analysis. In particular, we divide these techniques into three categories based on the geometric primitives they use to model facade features: points, rectangles, and polygons.

\paragraph{Pixel-wise segmentation}
Several works have focused on pixel-wise segmentation for facade analysis. For instance, \cite{schmitz_convolutional_2016} used cut-out images of facades to perform segmentation of windows and doors using convolutional neural networks (CNN). Similarly, \cite{liu2020deepfacade} proposed a method for semantic segmentation of windows and doors, introducing a symmetric loss function, enforcing most window predictions to be rectangular.
\cite{zhang_deep_2022} proposed a hierarchical deep learning framework that integrates several deep neural networks (PSPNet \citep{zhao2017pyramid}, DANet \citep{fu2019dual}, and DETR \citep{carion2020end}) for facade element detection to automatically detect various facade elements. Additionally, \cite{LU2023113275} used ResNet and BiFPN for instance segmentation, focusing on window detection and WWR calculation.
\cite{wang_improving_2024} developed a Vision Transformer (ViT)-based semantic segmentation method combined with line detection to improve facade element detection, capturing the shapes of buildings, windows, doors, roofs, and other elements.

\paragraph{Bounding box detection}
Despite providing valuable information, pixel-level segmentation is often an intermediate step because it lacks the geometric definition of facade elements typically required for subsequent analysis. In facade analysis, elements are commonly parameterized by rectangular bounding boxes.
\cite{teboul_parsing_2013} utilized reinforcement learning and Markov decision processes for bounding box detection from orthographic images. Similarly, \cite{mathias_atlas_2016} evaluated various methods for segmentation and object detection, using Conditional Random Fields (CRF) for classification and applying weak architectural principles for optimization to produce bounding boxes.
\cite{neuhausen_automatic_2018} proposed a method for converting perspective images to orthographic views, followed by detection using a soft cascaded classifier and post-processing to refine detections. This method assumed that windows are usually uniformly distributed within their rows or columns.
\cite{kong_enhanced_2021} focused on wall segmentation and window bounding box detection, converting wall segmentations into bounding boxes across multiple datasets. This approach highlighted challenges with bounding box accuracy due to perspective camera distortions.
\cite{rahmani_high_2018} utilized Structured Random Forest (SRF) and a Region Proposal Network (RPN) based on a CNN for asset bounding boxes in orthographic images, while \cite{szczesniak_method_2022} used computer vision techniques and GSV images to detect windows and calculate the WWR.
\paragraph{Polygon Detection}
Polygon detection methods are designed to capture the exact shape of facade elements. \cite{li_window_2020} developed a method for detecting keypoints of windows from perspective images, learning keypoint relationships to group them into window polygons.
\cite{cao_facade_2017} worked on detecting windows in aerial images using Hough forest classification and a proposed refinement of Hough voting for multiple object instances detection. Although this method is primarily designed for aerial images, its refinement approach is noteworthy.
\cite{wang_hierarchical_2024} proposed a method to detect windows and other elements in CAD drawings.
Furthermore, \cite{liu2020deepfacade} and \cite{ayala_deep_2021} explored orthorectification and facade segmentation and polygon detection from street view and satellite images, respectively.

\paragraph{Choosing a parameterization for downstream 3D modeling}
Selecting a suitable parameterization is critical for 3D modeling, balancing expressiveness and ease of use for downstream applications. Pixel-wise segmentation offers the highest flexibility, capturing detailed and complex shapes. However, it is difficult to use for downstream tasks because each pixel must be processed individually, leading to increased complexity in handling and storing the data. Polygon-based approaches provide a middle ground, capable of representing various shapes with fewer parameters than pixel-wise methods, but still requiring more complex handling. Rectangular bounding boxes, while the least flexible, offer the simplest and most user-friendly parameterization. They are typically sufficient for modeling most facade features and are easier to manage and integrate into downstream applications. It is worth noting that bounding box detection is more accurate in orthographic images, as this eliminates distortions present in perspective images, thereby enhancing the accuracy of 3D projections.

\begin{figure}[ht!]
\centering
\includegraphics[width=1\linewidth]{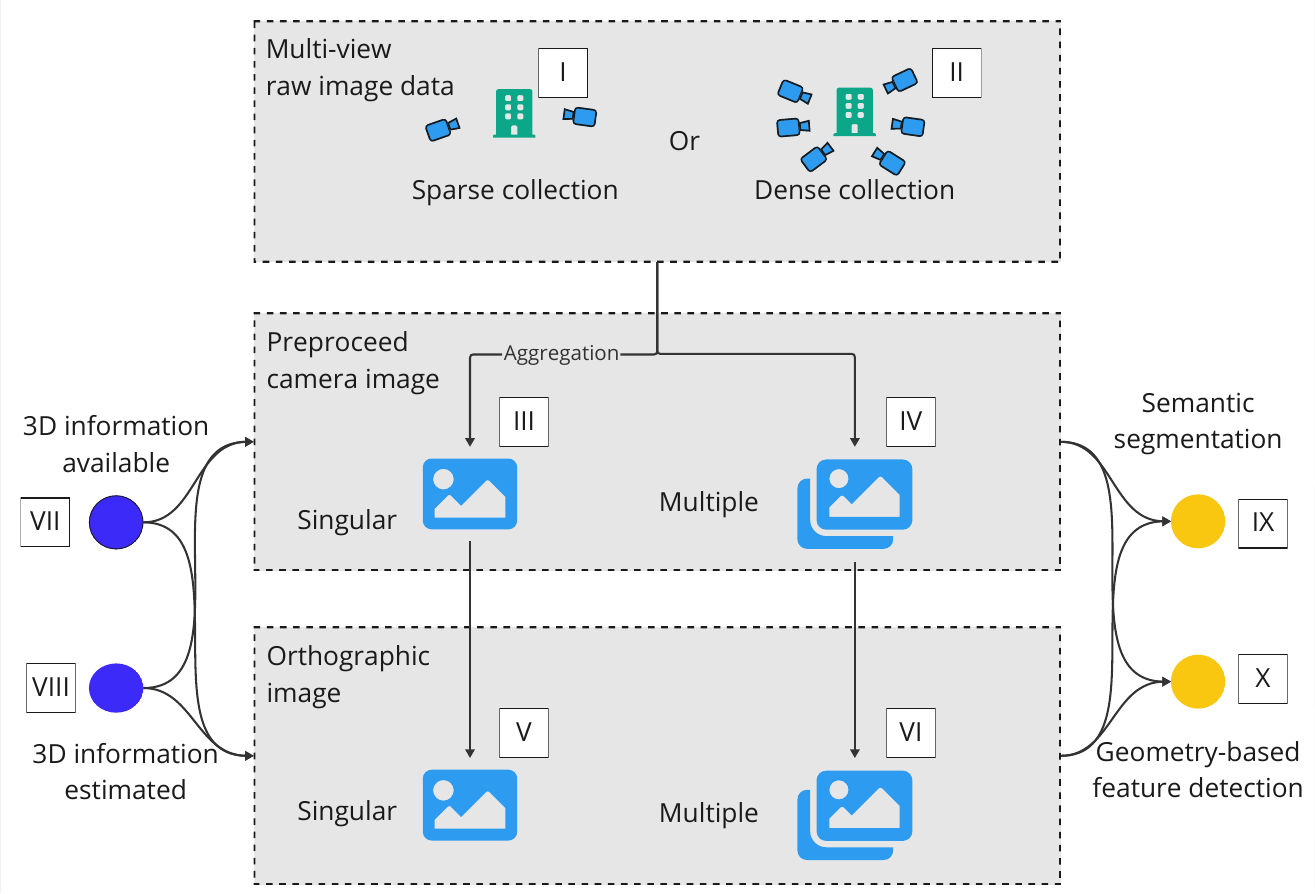}
\caption{\label{fig:literature-camera} Typical pipelines with camera images as primary input.}
\end{figure}

\subsubsection{3D modeling using camera images as the primary sensor}

Typical pipelines for facade analysis using camera images as the primary input are illustrated in Figure~\ref{fig:literature-camera}. The process begins with raw data collection, which can vary in density. Sparse collection methods (I) include large-scale drive-by street-level image acquisition, such as GSV, while dense collection (II) methods involve dense image captures of specific areas or buildings. These multi-view images can be used for estimating unstructured 3D information (such as point clouds) or parameterized 3D models (such as planes) on LoD2.

After the initial data collection, preprocessing steps are employed, including correcting for camera lens distortion and other image quality adjustments. Following preprocessing, the images can either be aggregated into a single composite image by fusing multiple views or selecting the best images based on certain criteria or treated as individual images. These two approaches correspond to components III and IV in the pipeline, respectively.
One additional step can be applied to correct perspective distortions in the images before semantic analysis on the facade.
Perspective distortions refer to the phenomenon where parallel lines appear convergent and distant objects appear smaller. These distortions can be problematic where the geometries in the image plane need to be maintained when transformed into 3D.
To address this, orthographic transformation (also referred to as orthorectification) can be applied to the images (V and VI). This transformation corrects perspective distortions, ensuring that parallel lines remain parallel and objects retain their consistent size regardless of their distance from the camera or their position in the image.
To establish the correspondence between pixels in the image plane and their 3D location, additional information is necessary such as 3D models or building blueprints. If 3D information is available (VII), it can be directly applied to project each pixel to a 3D location. Otherwise, this information must be estimated (VIII) from the images.

 Street-level imagery is commonly used for facade analysis and 3D modeling. These images capture static built environment features, pedestrians, cyclists, and vehicles, making them indispensable for various applications \cite{Goel2018}. Several web-based providers offer street imagery, with GSV and Bing StreetSide being among the most prominent \cite{Long2017}. GSV, in particular, stands out as a vast online browsable dataset consisting of billions of georeferenced street-level panoramic images from around the world. Since its inception in 2007, GSV has continuously updated its global image database, capturing panoramas every 5-10 meters in urban environments \cite{Cavallo2015}. Due to their large scale, these datasets are typically sparse around each individual building facade.

Some services, such as GSV API, also allows access to depth maps for specific panoramas, which can be decoded and visualized to reconstruct 3D planes. When these planes are not available, they can be estimated from the images using Structure-from-Motion (SfM) \citep{Hartley2004}. SfM creates 3D models from images by matching keypoints across multiple images, resulting in a 3D point cloud where each point is assigned a color based on the corresponding pixel value. When the images are not georeferenced, meaning that the camera poses are not available, SfM can estimate the poses. One can fit planes to these point clouds to obtain parameterized geometries subsequent analysis. While SfM is an effective method for 3D reconstruction, it requires dense image collection for effective keypoint matching.

When the data collection is dense, Neural Radiance Fields (NeRF) \citep{mildenhall2021nerf} is a relevant technique for 3D facade rendering. NeRF takes multi-view images and camera poses (measured or estimated from SfM) as input and achieves highly realistic rendering and novel viewpoint synthesis by training a neural network to predict color and density values along rays passing through a scene, effectively reconstructing complex lighting effects and details. To make the algorithm more accessible, \cite{mueller2022instant} provided tools with enhanced hyperparameter selection and optimization through real-time visualization of the rendering process. Furthermore, \cite{tancik2022block} demonstrated that NeRF could be applied to large-scale city rendering using street-level multi-view images.

Given these underlying image-to-3D techniques, there are several notable studies in the literature. \cite{fleet_learning_2014} introduced a method that uses multiple images for SfM to create a 3D mesh, focusing on segmenting and labeling the mesh from a single chosen image.
\cite{lotte_3d_2018} utilized multiple images to generate a mesh via SfM. They further applied a segmentation model to project segmentations onto the mesh, using overlapping images to determine the most frequent label for more accurate segmentation.
\cite{malihi_window_2018} leveraged drone (UAV) images to create a point cloud via SfM, which was georeferenced using ground control points. They implemented a color-based two-step selection process to detect window patches in the point cloud.
\cite{bacharidis_3d_2020} transitioned from stereo \citep{bacharidis2018fusing} to single mono camera images for their 3D reconstruction. They utilized deep learning for depth estimation to create a 3D point cloud, followed by segmentation of architectural features like windows and doors which were then assigned to the 3D point cloud.
Further, \cite{PAL2024105157} monitor construction activity progress, generating orthographic views via single-view projection when a suitable camera view exists, or falling back to NeRF rendering when single views are inadequate (e.g., due to face size, proximity, or occlusion). Semantic segmentation on these views yields area-based completion percentages. Compared to the orthographic representation based on virtual camera traversal presented in \cite{PAL2024105157}, our approach leverages NeRF to compute true orthographic projections directly from surface geometry. While camera traversal could potentially provide high-quality renders, our design is particularly suited for residential facades with long continuous sides, where traversal approaches may fail due to limited field of view, increased risk of occlusions, and inconsistencies in scale and alignment across multiple renderings. 

\cite{nishida_procedural_2018} took a different approach by using a single image with known building silhouettes in the image as inputs. Their method employs Convolutional Neural Networks (CNNs) to select a ``building mass grammar'' and estimate parameters, thereby determining the 3D structure of the building. This structure is used to create orthorectified facade images and to generate detailed 3D models including facade and window grammars.
\cite{pantoja-rosero_generating_2022} focused on using multiple images for SfM to estimate planes. They combined segmentation with projecting segmentations to the closest planes, refining these masks to generate bounding boxes on 3D planes for enhanced accuracy in 3D modeling.
\cite{ward_estimating_2023} employed images captured from a moving vehicle equipped with GNSS/IMU for precise camera localization and measuring poses. Their process involved SfM to build a 3D mesh, masking buildings, performing semantic segmentation, and projecting these masks onto the mesh. The mesh was rotated to fit the x-y plane, and bounding boxes were used for facade features, also estimating the building age from the images.
\cite{salehitangrizi_3d_2024} proposed a dual-track method using multi-view images and building footprints. They utilized Faster R-CNN and the Segment Anything model \citep{kirillov2023segment} to project 2D borders into 3D using pinhole camera models and collinearity equations. LiDAR scanner data can also integrated into this workflow.

\cite{zhang_slod2win_2024} utilized a 3D model as input to generate orthogonal facade images. They employed Faster R-CNN for bounding box detection and clustering-based window alignment to ensure consistency in window dimensions and positions. This method also included glass plane detection to enhance window detail and model floors and ceilings.
\cite{tarkhan_facade_2024} uses a grammar-based edge detection framework and a learning-based method utilizing CNNs and compares the result in New York and Lisbon. The paper finds that the learning-based method generally performs better, and proposes a hybrid approach to leverage strengths of both methods.
In addition, \cite{wong_semantic_2024} focused on creating an extensive instance segmentation dataset for interior and exterior scenes, aimed at 3D reconstruction. The dataset, aligned with standardized data schemas like Industry Foundation Classes (IFC), ensures accurate geometric representations and topological relationships in segmentation results.

In summary, these studies collectively highlight the evolution of 3D reconstruction techniques from 2D images, incorporating advanced segmentation methods, deep learning, and multi-view geometry to achieve increasingly accurate and detailed models. A summary of the aforementioned techniques can be found in Table~\ref{tab:summary}.

\subsubsection{3D modeling using both camera and LiDAR as the primary sensors}

Although less relevant to our paper, as we primarily use cameras, point processing is still worth mentioning since SfM produces point clouds, making some techniques transferable. We include these existing works for completeness.

\cite{dehbi_statistical_2017} explore the learning of weighted attributed context-free grammar rules for 3D building reconstruction.
They employ an Support Vector Machine (SVM) for the classification of facade structures and an Multi-Layer Neural Network (MLN) for estimating parameters of facade parts.
\cite{li_hierarchical_2017} present a hierarchical approach to facade point cloud analysis. Their method does not incorporate color information. Instead, they segment the point cloud into "principal facade planes" and "2.5D segments" (images with depth). The BieS algorithm extends the samples, and the ScSPM algorithm extracts features. A linear SVM classifier then categorizes each superpixel into semantic classes such as window, wall, roof, shop, or door based on the learned features.
\cite{fryskowska_no-reference_2018} focus on the creation of 3D models of cultural heritage buildings using RGB point clouds.
Their work emphasizes the generation of 3D models without segmenting the point cloud. The models are created using Autodesk Revit software, with comparisons made between manual and automatic methods and the point cloud.
\cite{gadde_efficient_2018} use both ortho and projection images or point clouds as input for segmentation. They apply boosting decision trees to segment the images and point clouds separately and evaluate their performance both individually and in combination. However, the paper does not address the challenges of combining point clouds with camera data.
\cite{wen_accurate_2019} focus on merging airborne and terrestrial point clouds, filtering out ground points, and extracting planes using RANSAC. They utilize oblique images taken from the sky to find the best angle for each plane, projecting the images onto the planes. The plane edges are aligned with line detections in the images to create a colored 3D model, which is manually edited to segment windows and other features.
\cite{fan_layout_2021} use point clouds as input and apply RANSAC to segment balconies and windows. They construct a hierarchical graph representing the facade, with levels for the facade, floor, and window. Repeating objects are clustered in the graph based on their similar size and spacing.

\subsubsection{Research gap and differences to our approach}
Our methodology differs from existing techniques by avoiding the conventional process of ``pixel-wise segmentation in image $\rightarrow$ projecting these labels to 3D points (derived from images) $\rightarrow$ defining geometry in 3D from these points.''
This traditional approach is challenging for downstream thermal modeling due to the difficulty in estimating window parameters in 3D when the point cloud derived from images is sparse and the segmentation results are erroneous. Instead, we estimate geometric primitives in the image plane. Specifically, we generate an orthographic image where perspective distortions are corrected to enhance accurate shapes and sizes. We then perform bounding box detection directly on this orthographic image. This method leverages the capability of semantic analysis and geometry parameterization in images (as opposed to sparse point clouds), leading to a more robust estimation of window geometry.

Note that orthographic transformation is crucial for estimating accurate geometry on the facade plane. Additionally, we utilize orthographic images as a unified interface for both sparse and dense data collection, enhancing usability by accommodating varying levels of data availability.
SI3FP performs best when the feature surfaces parallel to the facade planes are flat, which is a realistic assumption for our use case.

For sparse data collection, most literature focuses on selecting the best single perspective image for detecting building facades. However, this approach is insufficient due to severe occlusion and perspective distortion that can occur even in the best images. Instead, we developed an ensemble method that aggregates all available images to enhance robustness and accuracy. This aggregation helps mitigate the impact of occlusions by combining information from multiple viewpoints.

Moreover, for dense data collection, there is a noticeable gap in research concerning the application of NeRF to building facade parsing. \cite{hachisuka2023harbingers} conducted a case study on semantic segmentation of building structures using 3D point clouds. In contrast, our work applies NeRF to create a detailed photorealistic 3D render of buildings, followed by an orthographic transformation of the facade.
By using NeRF, we achieve higher-quality reconstructions compared to traditional Multi-View Stereo (MVS) methods \citep{schoenberger2016mvs}, particularly in handling complex lighting conditions and reflections common in urban environments.
These images are subsequently used for window detection, employing pre-trained deep learning models. This approach simplifies the facade parsing process and enhances the reliability of the geometric estimation of building facades.

\section{Method}
\label{sec:method}

\begin{figure*}[ht!]
\centering
\includegraphics[width=0.75\linewidth]{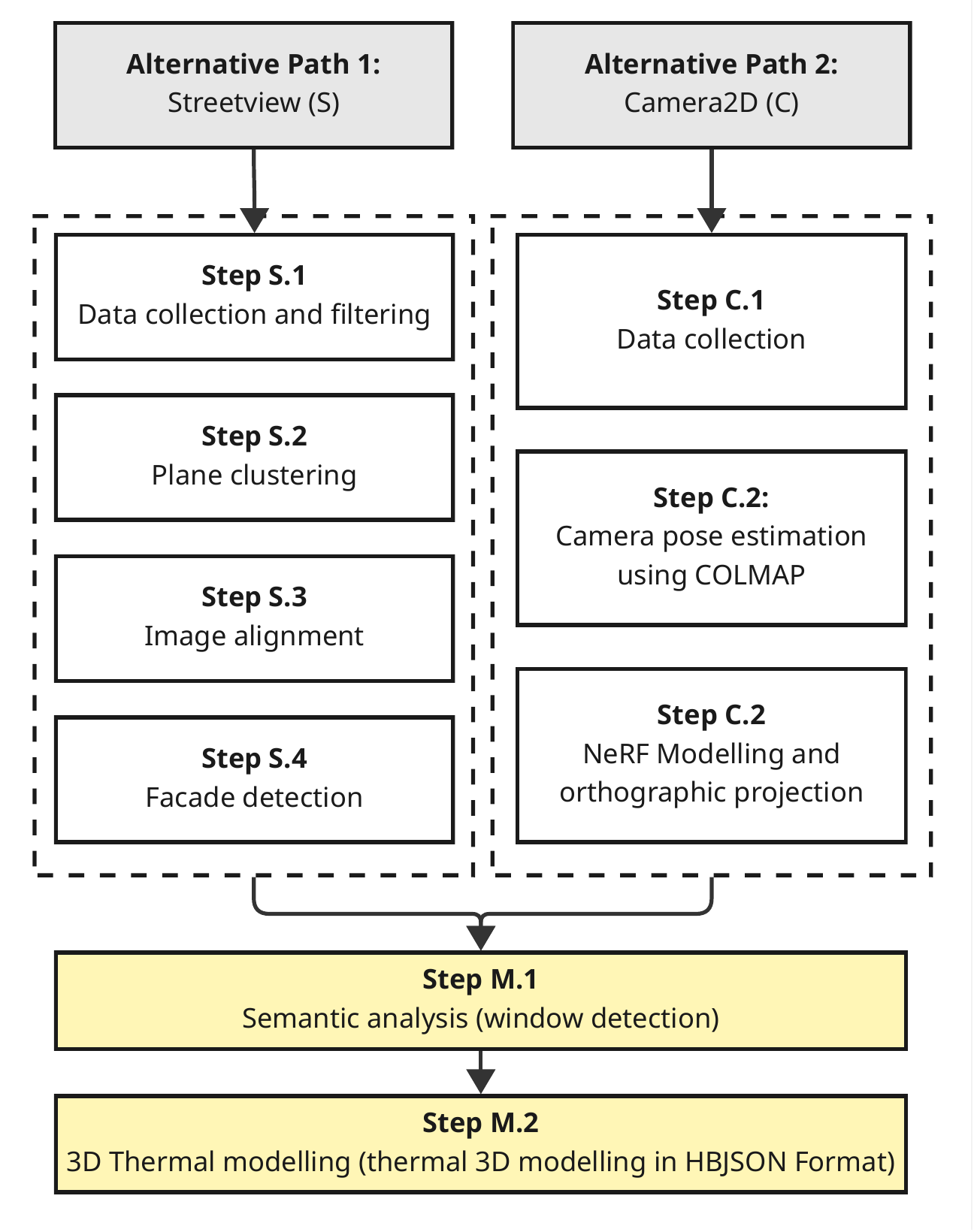}
\caption{\label{fig:overview}  Overview of SI3FP, which uses camera images as input to generate true-to-scale orthographic images for semantic façade parsing and 3D thermal modeling.}
\end{figure*}

The SI3FP pipeline consists of two alternative paths offering their respective trade-offs: 1) the \emph{StreetView} path for scalable inspection, and
2) the \emph{Camera2D} path for targeted inspection.
An overview of these two paths can be found in Figure~\ref{fig:overview}.
Each path has its individual data collection and processing steps: four steps (S.1-S.4) for StreetView and three steps (C.1-C.3) for Camera2D. They then converge into two merged steps, denoted as M.1 and M.2 for semantic facade parsing and thermal modeling.
\subsection{StreetView (S) for scalable inspection}\label{sec:streeview_pipeline}

\begin{figure*}[ht!]
\centering
\includegraphics[width=1\linewidth]{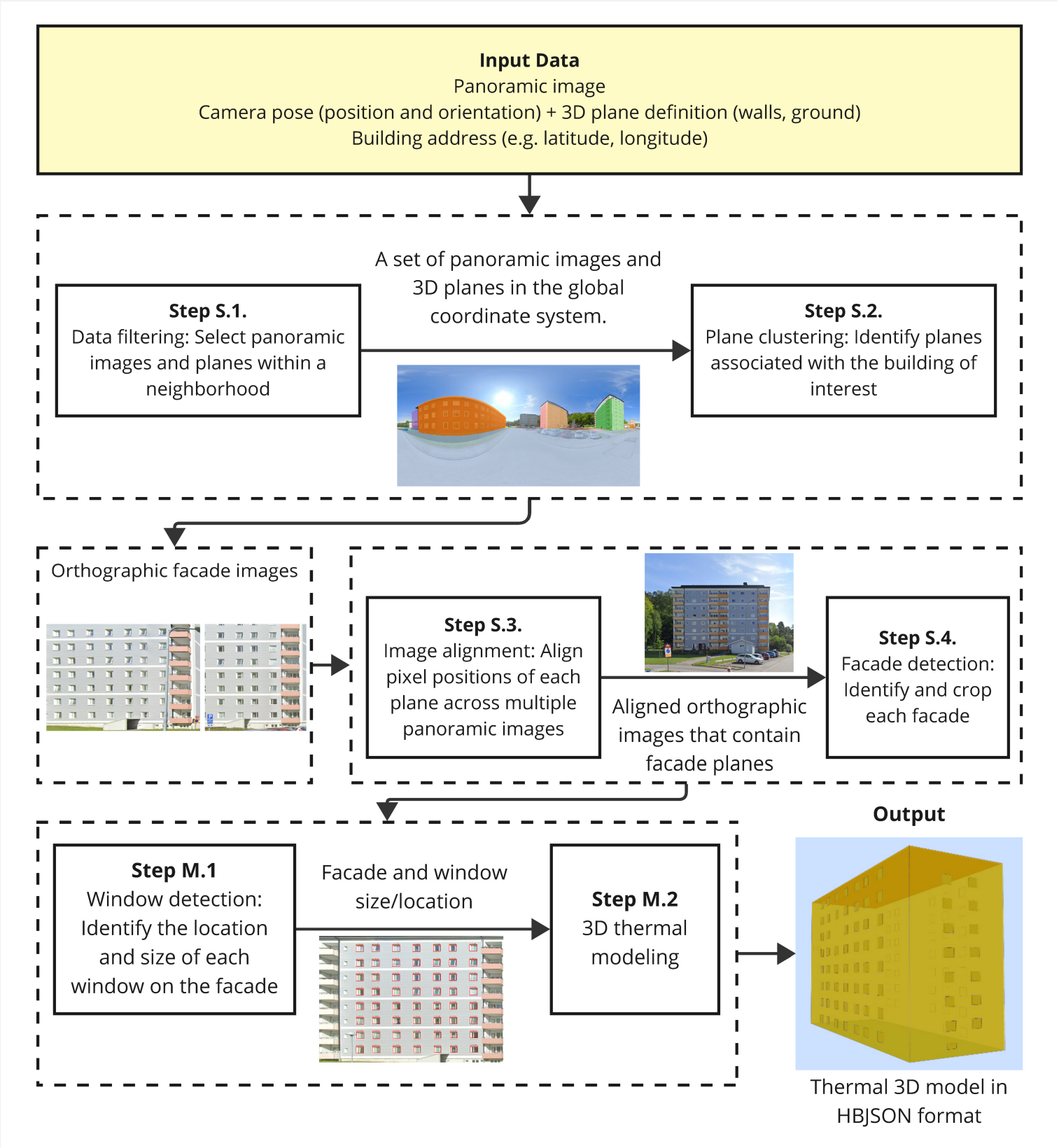}
\caption{\label{fig:workflow-streetview} Workflow for the StreetView path.}
\end{figure*}
Street level imagery can be obtained through multiple methods. By employing a mobile data collection platform, such as a vehicle or drone, equipped with video cameras, a gyroscope, distance sensor (such as a LiDAR or a radar system), and a location sensor (e.g. GNSS receivers), data can be captured across diverse environments at scale. Alternatively, access to large-scale street view datasets is possible through specialized services dedicated to offering extensive street view imagery.
Although our focus is on data collection via Google's Street View (GSV) API, the outlined process can be adapted for use with other data sources or collection methods.
An overview can be found in Figure~\ref{fig:workflow-streetview}.
The StreetView path consists of five steps described below (Step S.1-S.4).

\begin{figure}[ht!]
\centering
\includegraphics[width=0.8\linewidth]{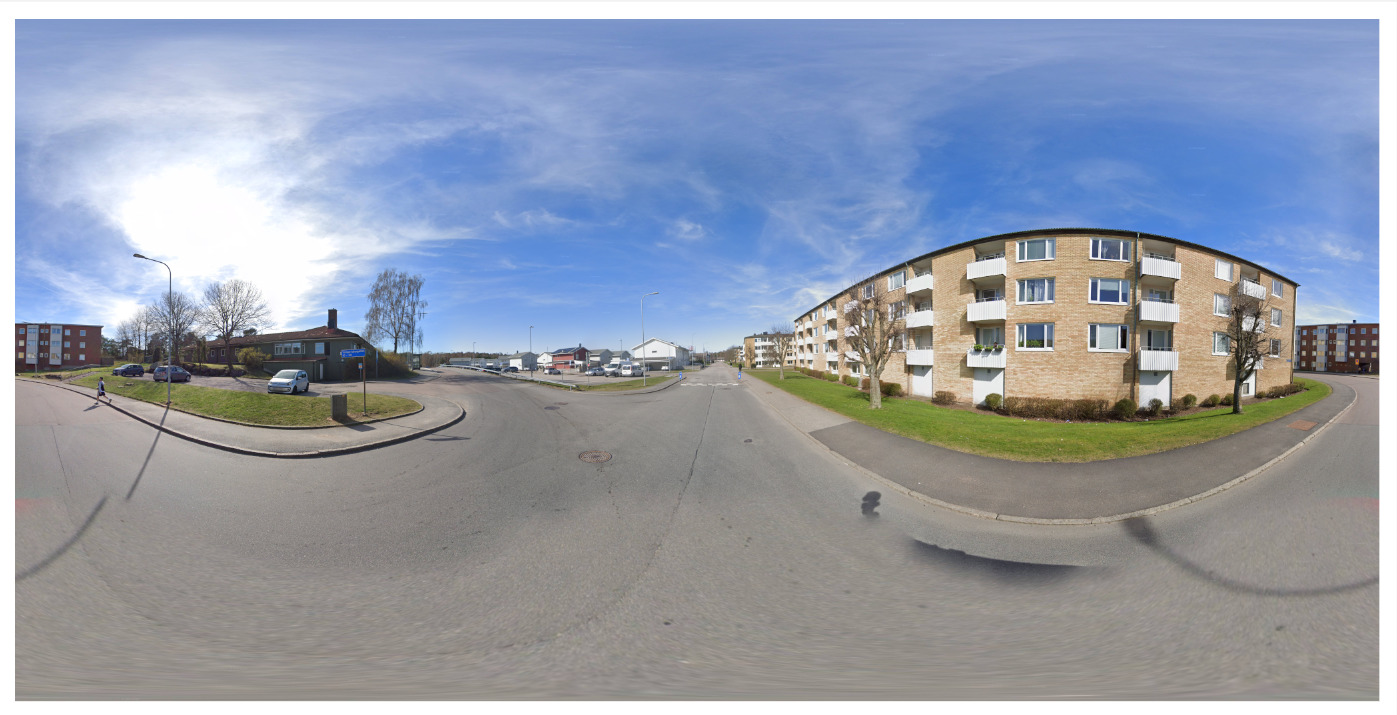}
\caption{Example of a 16384 $\times$ 8192 panoramic image captured at a given location.}\label{fig:streetviewpano}
\end{figure}

\paragraph{Step S.1 Data collection and filtering:} Select panoramic images and meta data.

\begin{figure}[ht!]
\centering
\includegraphics[width=1\linewidth]{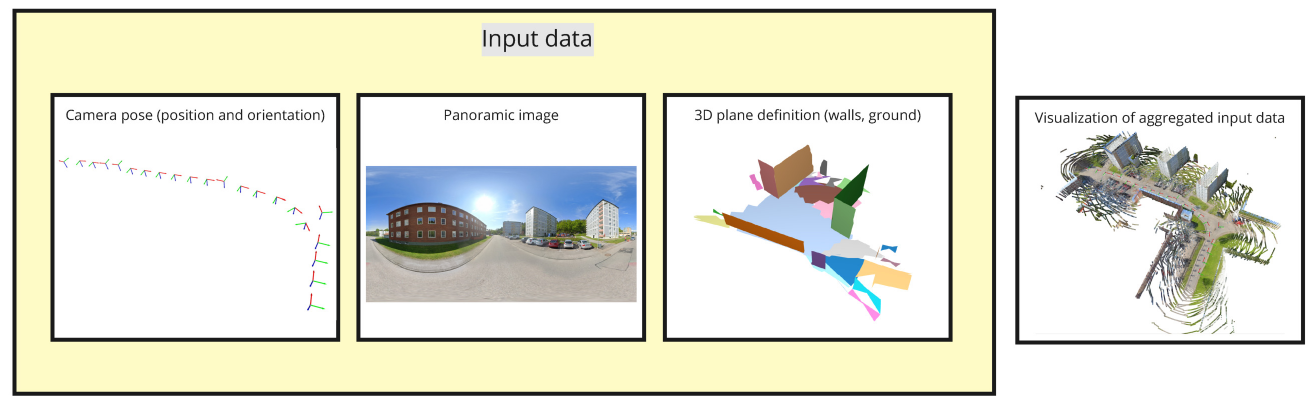}
\caption{\label{fig:inputs-streetview} Input data for StreetView. }
\end{figure}

The complete input data for StreetView can be visualized in Figure~\ref{fig:inputs-streetview}.
The initial step involves selecting panoramic images. While an important use case is to collect these images continuously across various locations, for the sake of simplicity, we focus on illustrating the data collection method within a specific neighborhood that is identified by a central geographic coordinate (marked by its latitude and longitude) and a radius.
Utilizing GSV API, we can pinpoint nearby panoramic views identified by a unique PanoID for the given geographical coordinates.
Each PanoID represents a 360-degree panoramic image, providing detailed meta data such as latitude, longitude, altitude, heading (direction), pitch (angle of elevation), roll (axial tilt), capture date, and connections to adjacent views.
The selection of the \emph{origin view}, or starting point, is achieved by opting for the most recently captured image among the search results for the given location. This enables the exploration of various street view locations throughout the targeted neighborhood by traversing the links to neighboring views.
When there are sufficiently many images available for one building, a filtering criterion is applied to include only those images captured on the same date to maintain consistency across the collected images (e.g. uniform lighting, contrast, and scene composition, among other environmental conditions).
The outcome of this step is a set of panoramic images\footnote{The size of each panoramic image provided by the current version is 16384 $\times$ 8192 pixels.}(cf. Figure~\ref{fig:streetviewpano}) with their respective meta data.

In addition to panoramic images, GSV provides plane definitions. More specifically, walls and ground surfaces are represented as planes in the world coordinate system.
The information is typically gathered by a sensor that provides 3D information (e.g. a stereo camera or a LiDAR scanner), and it can be refined by retrieving the building locations and from databases of building locations and outlines offered by various official or unofficial organizations (sometimes referred to as the land registry).

Each panoramic image is associated with a camera pose. Each pixel in the panoramic image can be associated with a plane, allowing for transformation of the pixel value between the camera coordinate system and the world coordinate system in 3D.
This association, if not readily given, can be estimated by standard computer vision techniques \citep{Hartley2004}. More precisely, to map a pixel from a panoramic image directly to a 3D point on a specified plane, we first convert pixel coordinates \((x, y)\) to spherical angles \(\theta\) and \(\phi\), and then to a unit direction vector \(\mathbf{V}\). The intersection of \(\mathbf{V}\) with the plane defined by normal vector \([a, b, c]\) and distance \(d\) is determined by scaling \(\mathbf{V}\) by \(\frac{d}{aV_x + bV_y + cV_z}\). This scaled \(\mathbf{V}\) gives us the 3D coordinates.

\paragraph{Step S.2 Plane clustering:} Identify planes associated with the building of interest.

\begin{figure}[h!]
\centering
\includegraphics[width=0.8\linewidth]{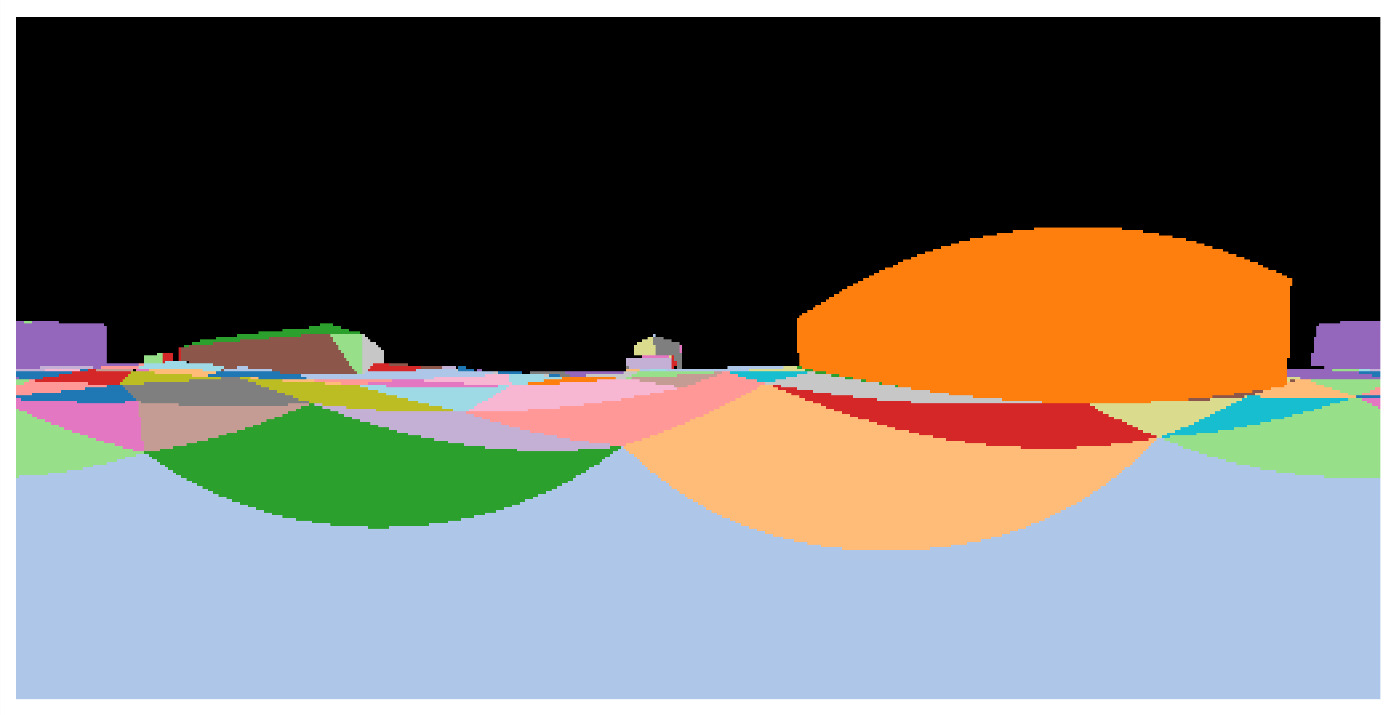}
\caption{A $512\times 256$ matrix showing the plane index (plane$_{\text{id}}$) associated with the pixel values in the panoramic image.}\label{fig:streetviewplaneindex}
\end{figure}

To extract 3D information and enable detailed analysis, each RGB panoramic image captured must be connected to a 3D framework.
Particularly within GSV imagery, every panoramic image is linked to a set of 3D planes.
In detail, for each panoramic image, there exists a plane association matrix sized
$512\times 256$. Each element within this matrix corresponds to a specific plane. The alignment of this matrix is in sync with the panoramic image from which it is derived, although it is a downsampled version relative to the original image's resolution ($16384 \times 8192$). This can be visualized in Figure~\ref{fig:streetviewplaneindex}. This is specific to how GSV defines the association between the panoramic image and their corresponding 3D structure.
It is noteworthy that some pixels may lack a plane association. This can be caused by the limitation of the data collection equipment and process.

\begin{figure}[h!]
\centering
\includegraphics[width=0.8\linewidth]{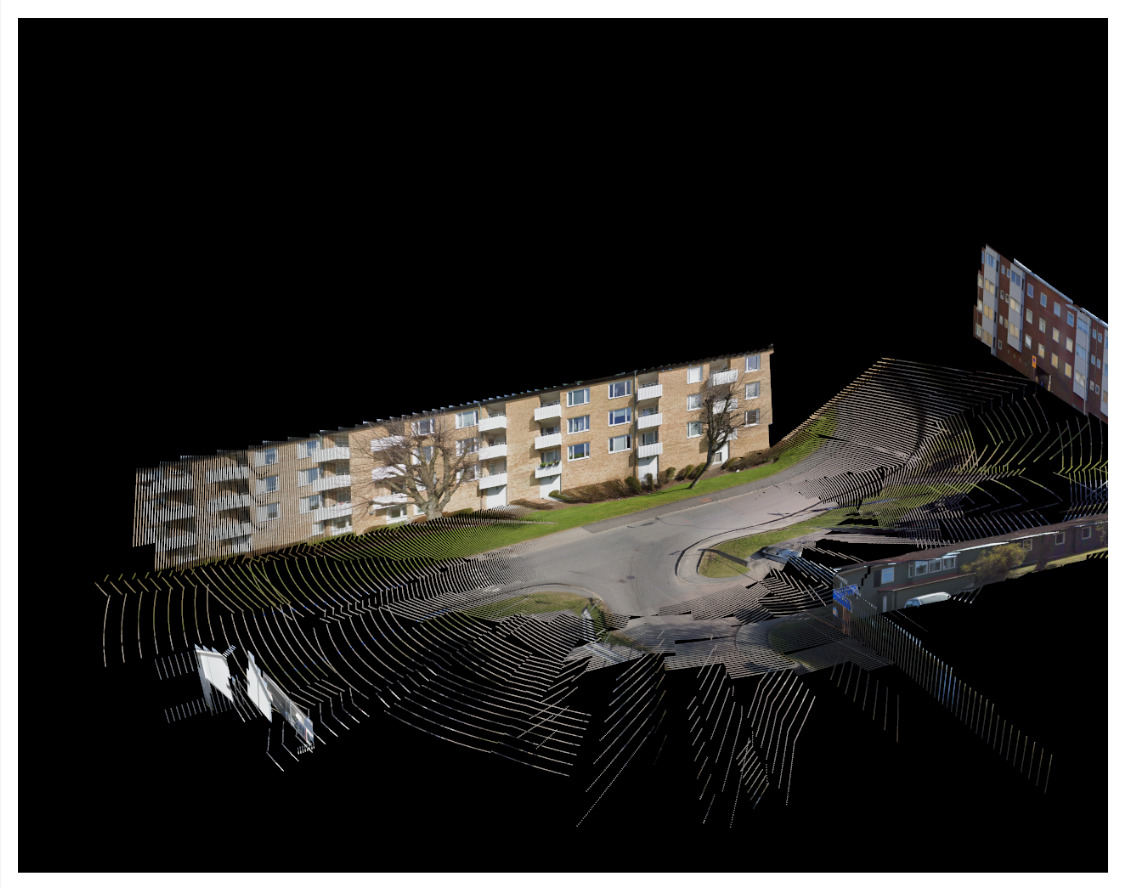}
\caption{\label{fig:streetview3d} Point cloud generated by projecting the panoramic image pixel values onto their corresponding 3D location.}
\end{figure}

With the plane association matrix, in combination with the planes, the RGB data from the panoramic images can be mapped onto these planes, infusing the 3D structure with color and texture. This step results in a colored point cloud in 3D (cf. Figure~\ref{fig:streetview3d}).

\begin{figure}[h!]
\centering
\includegraphics[width=1\linewidth]{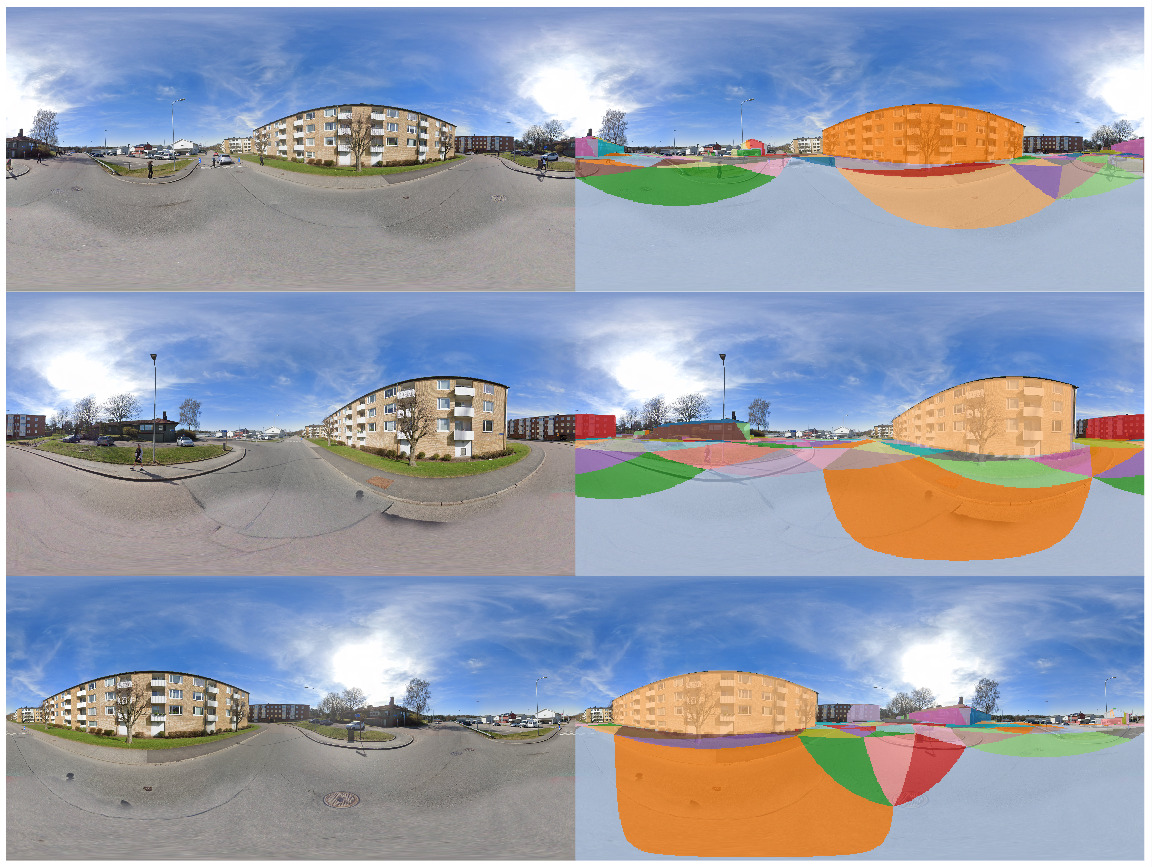}
\caption{\label{fig:streetviewmultiplepano} Multiple, nearly potentially overlapping panoramic images along with their plane association matrices.}
\end{figure}

To enhance the robustness of the system, we repeat the aforementioned process for all nearby, potentially overlapping panoramic images collected in Step S.1 (cf. Figure~\ref{fig:streetviewmultiplepano}). Every panoramic image contributes its unique planes to the collective model. We then adjust these planes into a common coordinate framework through translation and rotation.

The next critical step is to cluster these transformed 3D planes from different panoramas to identify those corresponding to the same physical facade surface. We use the Agglomerative Clustering algorithm with average linkage.
The distance $D(i, j)$ between any two plane segments $i$ (unit normal $\mathbf{n}_i$, origin distance $d_i$) and $j$ ($\mathbf{n}_j, d_j$) is computed using a custom metric:
\begin{equation} 
\label{eq:plane_dist}
    D(i, j) = (1 - \mathbf{n}_i \cdot \mathbf{n}_j) + 0.01 \cdot |d_i - d_j|
\end{equation}
This metric combines the cosine distance between normals $(1 - \mathbf{n}_i \cdot \mathbf{n}_j)$, weighting orientation similarity highly ($w_n=1.0$), with the weighted absolute difference in origin distances $|d_i - d_j|$ ($w_d = 0.01$). A strict distance threshold of $1 \times 10^{-5}$ is used for clustering.

Following clustering, orthographic views are generated from each contributing Street View panorama in a cluster. A common 3D plane and coordinate system are established for the cluster. This process is summarized in Algorithm~\ref{alg:per_pano_ortho_math}.

An orthographic image grid is defined on this plane at a desired real-world resolution (e.g., pixels per meter). For each grid pixel, the corresponding 3D point on the common plane is projected back into the source panoramic image using its known pose and geometry to determine the source pixel coordinates. The color is then sampled from the panorama and assigned to the orthographic grid pixel. This is the orthographic transformation step, described in Algorithm~\ref{alg:streetview-ortho}.

This per-panorama process yields multiple, consistently scaled orthographic views of the same facade area. These views effectively correct the perspective distortion of the original panoramas, ensuring parallel lines remain parallel, preserving scale, and reducing parallax errors. They serve as the input for the image alignment in Step S.3.
\begin{algorithm}[htbp] 
\scriptsize 
\caption{StreetView-based Orthographic Projection}\label{alg:streetview-ortho}
\begin{algorithmic}[1] 
    \State \textbf{Input:} Panoramic image $\mat{I}_{pano}$, Plane equation $\text{Eq}_{plane}$ ($\vect{n}_{plane}, d$), 3D points $\set{P}_{plane3D}$, Pixel size $s_{pixel}$, Panorama pose $\mat{T}_{pano}$ (implicit)
    \State \textbf{Output:} Orthographic image $\mat{I}_{ortho}$

    \Statex \textbf{Stage 1: Define Plane Geometry \& Coordinate System}
    \State $\vect{n}_{plane} \gets \text{ExtractNormal}(\text{Eq}_{plane})$
    \State $\vect{u}_{plane}, \vect{v}_{plane} \gets \text{CalculatePlaneBasisVectors}(\vect{n}_{plane})$
    \State $\mat{R}_{w2p} \gets \text{RotationMatrixFromBasis}(\vect{u}_{plane}, \vect{v}_{plane}, \vect{n}_{plane})$

    \Statex \textbf{Stage 2: Determine Extent and Output Grid}
    \State $\set{P}_{plane2D} \gets \text{TransformToPlaneCoords}(\set{P}_{plane3D}, \mat{R}_{w2p}, \text{Eq}_{plane})$
    \State $\vect{p}_{min2D}, \vect{p}_{max2D} \gets \text{CalculateExtent}(\set{P}_{plane2D})$
    \State $W_{out} \gets \text{Round}((\vect{p}_{max2D}.x - \vect{p}_{min2D}.x) / s_{pixel})$
    \State $H_{out} \gets \text{Round}((\vect{p}_{max2D}.y - \vect{p}_{min2D}.y) / s_{pixel})$
    \State $\mat{I}_{ortho} \gets \text{CreateImageBuffer}(W_{out}, H_{out})$
    \State $\vect{p}_{gridOrigin2D} \gets \vect{p}_{min2D}$

    \Statex \textbf{Stage 3: Generate Orthographic Image Pixels}
    \For{$y \gets 0$ to $H_{out} - 1$}
        \For{$x \gets 0$ to $W_{out} - 1$}
            \State $\vect{p}_{plane2D} \gets \vect{p}_{gridOrigin2D} + \text{Vector2D}(x \cdot s_{pixel}, y \cdot s_{pixel})$
            \State $\mat{R}_{p2w} \gets \text{Inverse}(\mat{R}_{w2p})$
            \State $\vect{p}_{world} \gets \text{TransformToWorldCoords}(\vect{p}_{plane2D}, \mat{R}_{p2w}, \text{Eq}_{plane})$
            \State \Comment{Core step: Reproject 3D point to 2D panorama}
            \State $u_{pano}, v_{pano} \gets \text{ProjectWorldToPano}(\vect{p}_{world}, \mat{T}_{pano}, \mat{I}_{pano}.\text{shape})$
            \If{$\text{IsValidCoord}(u_{pano}, v_{pano}, \mat{I}_{pano}.\text{shape})$}
                \State $C_{pixel} \gets \text{SampleColor}(\mat{I}_{pano}, u_{pano}, v_{pano})$
                \State $\mat{I}_{ortho}[y, x] \gets C_{pixel}$
            \Else
                \State $\mat{I}_{ortho}[y, x] \gets \text{BackgroundColor}$
            \EndIf
        \EndFor
    \EndFor

    \State \textbf{return} $\mat{I}_{ortho}$
\end{algorithmic}
\end{algorithm}

\begin{algorithm}[htbp] 
\scriptsize 
\caption{StreetView Geometric Plane Clustering}\label{alg:plane_clustering} 
\begin{algorithmic}[1] 
    \State \textbf{Input:} Dictionary $\set{P}$ containing panoramic data (incl. local planes and world transform $\mat{T}$), Clustering distance threshold $\delta_{cluster}$ ($= 1 \times 10^{-5}$)
    \State \textbf{Output:} Dictionary $\set{C}_{geom}$ mapping cluster IDs to lists of $(pano\_id, plane\_idx)$ tuples for geometrically similar plane segments.

    \Statex \textbf{Stage 1: Transform Candidate Planes to World Coordinates}
    \State $\set{P}_{world} \gets \text{empty list}$ \Comment{List to store world-frame plane equations $p_{world}=(\vect{n}, d)$}
    \State $\set{L}_{idx} \gets \text{empty list}$ \Comment{List to store corresponding $(pano\_id, plane\_idx)$}
    \For{$pano\_id \in \text{Keys}(\set{P})$}
        \For{$plane\_idx \gets 0$ to $\text{NumPlanes}(\set{P}[pano\_id]) - 1$}
            \State $p_{local} \gets \set{P}[pano\_id].\text{depth.planes}[plane\_idx]$
            \If{$\text{IsHorizontal}(p_{local})$ or $\text{IsZero}(p_{local})$} \Comment{Filter non-facade planes}
                \State \textbf{continue}
            \EndIf
            \State $\mat{T} \gets \set{P}[pano\_id].T$ \Comment{Get panorama's transform to world}
            \State $p_{world} \gets \text{TransformPlane}(p_{local}, \mat{T})$ \Comment{Calculate plane equation in world frame}
            \State $\set{P}_{world}.\text{append}(p_{world})$
            \State $\set{L}_{idx}.\text{append}((pano\_id, plane\_idx))$
        \EndFor
    \EndFor

    \Statex \textbf{Stage 2: Cluster Transformed Planes by Geometric Similarity}
    \State \Comment{Calculate pairwise distances using metric from Equation \ref{eq:plane_dist}}
    \State $\mat{D} \gets \text{CalculatePairwiseDistances}(\set{P}_{world}, \text{CustomMetric})$
    \State \Comment{Apply Agglomerative Clustering}
    \State $l_{geom} \gets \text{AgglomerativeClustering}(\mat{D}, \text{linkage='average'}, \text{threshold}=\delta_{cluster})$

    \Statex \textbf{Stage 3: Collect Cluster Members}
    \State $\set{C}_{geom} \gets \text{empty dictionary}$
    \For{$label \in \text{UniqueLabels}(l_{geom})$}
        \If{$label = -1$} \Comment{Skip potential noise label}
            \State \textbf{continue}
        \EndIf
        \State $idx_{members} \gets \text{IndicesWhere}(l_{geom} == label)$
        \If{$\text{Length}(idx_{members}) \ge 1$} \Comment{Keep clusters with at least one member}
          \State $\set{M} \gets [\set{L}_{idx}[i] \text{ for } i \in idx_{members}]$ \Comment{Get list of $(pano\_id, plane\_idx)$}
          \State $\set{C}_{geom}[label] \gets \set{M}$
        \EndIf
    \EndFor

    \Statex \Comment{Output $\set{C}_{geom}$ identifies groups of geometrically similar plane segments.}
    \State \textbf{return} $\set{C}_{geom}$
\end{algorithmic}
\end{algorithm}

\begin{algorithm}[htbp]
\scriptsize
\caption{StreetView Per-Segment Orthographic Image Generation}\label{alg:per_pano_ortho_math}
\begin{algorithmic}[1]
    \State \textbf{Input:} Clustered plane segments $\set{C}_{geom}$ (from Algorithm \ref{alg:plane_clustering}), Panorama data $\set{P}$, Pixel size $s_{pixel}$
    \State \textbf{Output:} Dictionary $\set{I}_{ortho}$ mapping each $(pano\_id, plane\_idx)$ to its orthographic image $\mat{I}_{ortho\_seg}$.
    \State $\set{I}_{ortho} \gets \text{empty dictionary}$
    \For{$clusterID \in \text{Keys}(\set{C}_{geom})$}
        \State $\set{M} \gets \set{C}_{geom}[clusterID]$ \Comment{List of members $(pano\_id, plane\_idx)$}
        \If{$\text{Length}(\set{M}) == 0$} \textbf{continue} \EndIf

        \State \textit{// Generate orthographic image for each segment in the cluster}
        \For{$(pano\_id, plane\_idx) \in \set{M}$}
            \State $\mat{I}_{pano} \gets \text{LoadPanoImage}(\set{P}[pano\_id])$
            \State $p_{local} \gets \set{P}[pano\_id].\text{depth.planes}[plane\_idx]$ \Comment{Segment's local plane equation}
            \State $\set{P}_{seg3D} \gets \text{ExtractPointsForSegment}(\set{P}[pano\_id], plane\_idx)$ \Comment{Segment's 3D points}
            \State $\mat{T}_{pano} \gets \set{P}[pano\_id].T$ \Comment{Segment's panorama pose}

            \If{$\set{P}_{seg3D}$ is empty} \textbf{continue} \EndIf

            \State \textit{// Generate ortho view using segment's data and panorama}
            \State \Comment{Applies the principle from Algorithm \ref{alg:streetview-ortho}}
            \State $\mat{I}_{ortho\_seg} \gets \text{GenerateSingleOrthoStreetView}(\mat{I}_{pano}, p_{local}, \mat{T}_{pano}, \set{P}_{seg3D}, s_{pixel})$

            \State $\set{I}_{ortho}[(pano\_id, plane\_idx)] \gets \mat{I}_{ortho\_seg}$
        \EndFor
    \EndFor

    \Statex \Comment{Output $\set{I}_{ortho}$ contains individual segment views, ready for alignment.}
    \State \textbf{return} $\set{I}_{ortho}$
\end{algorithmic}
\end{algorithm}

\paragraph{Step S.3 Image alignment:} Align pixel positions of each plane across multiple panoramic images.

\begin{figure}[h!]
\centering
\includegraphics[width=1\linewidth]{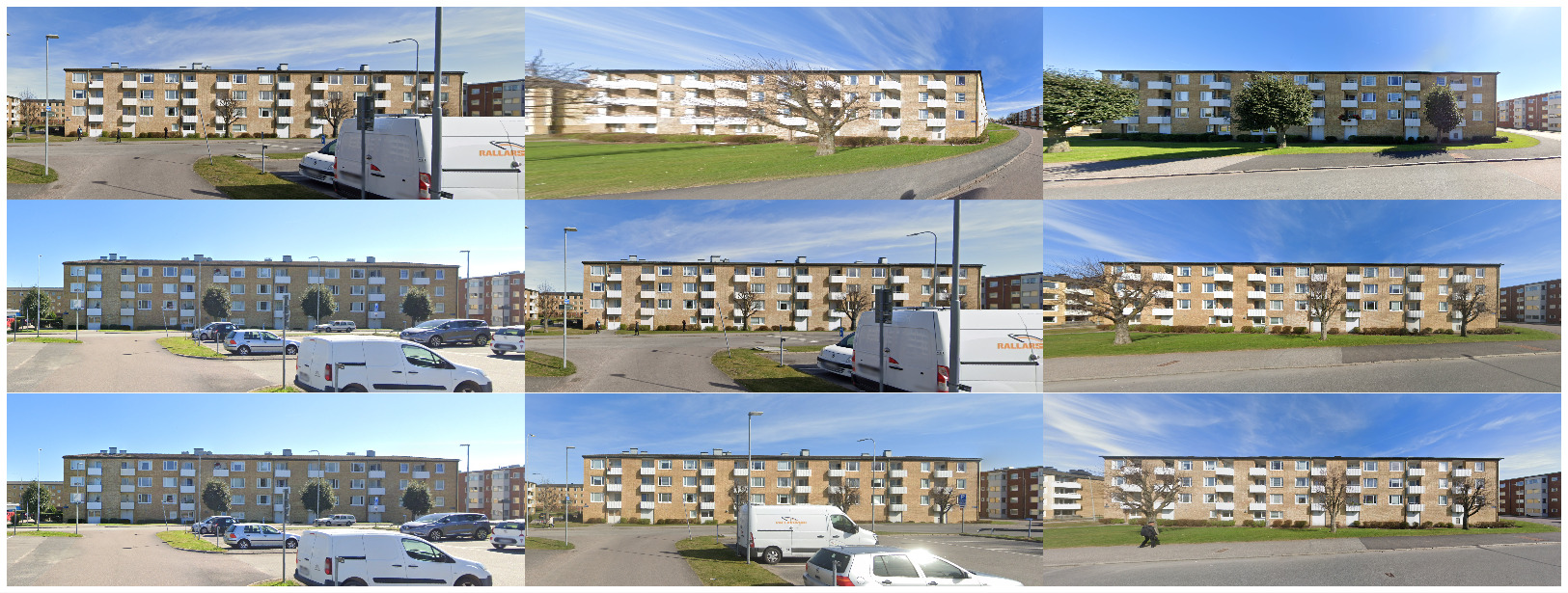}
\caption{\label{fig:streetviewaligned} Aligning orthographic images for improving robustness. We employ keypoint detection and alignment techniques to align the orthographic images. This alignment ensures that identical pixel locations across different images correspond to the same physical point in 3D space.}
\end{figure}

Challenges such as incorrect plane definitions, camera poses, and misalignments can arise. Therefore, we would like to ensemble results from different orthographic images to enhance the robustness of the subsequent semantic analysis. To achieve this, SIFT key points detection \citep{lowe2004distinctive} and image registration \citep{Hartley2004} are employed between each pair of panoramic images. It is worth noting that this process, aimed at aligning facades and windows, can be time-consuming due to its pairwise complexity. The outcome of this step is a collection of aligned images as shown in Figure~\ref{fig:streetviewaligned}.

\paragraph{Step S.4 Facade detection:} Detect and crop each facade.

The next objective is to extract the facades of interest from these orthographic images (cf.~\ref{fig:streetviewaligned}).
In theory, these facades can be extracted in the image based on the plane definition.
However, in reality, the plane definition is often not well aligned or complete when provided at scale.
This causes issues such as planes do not cover the full facade, incorrect facade size, etc.
Therefore, we extract the facade algorithmically with a bounding box.

We start by detecting lines and identify the vertical and horizontal ones by estimating their angles with a predefined tolerance of 10 degrees.
Among these lines, those that appear consistently across multiple images are considered reliable and categorized as \emph{relevant lines}, whereas all others are considered \emph{irrelevant lines}.
For each orthographic image, a RANSAC-inspired method \citep{fischler1981random} proposes candidate facade boundaries using lines from the reliable set (derived from consistent lines across views) and scores these boundaries based on how well all detected lines in the current image fit, implicitly favouring structurally sound alignments found within the reliable lines.
The algorithm is described in Algorithm~\ref{alg:ransac}. Following this identification, we crop the facade from the image.

Note that the effectiveness of this process relies on the presence of dominant horizontal and vertical lines on the facade, which allows the scoring mechanism to identify correct boundaries, though this characteristic might be less prevalent on irregular buildings.

\begin{algorithm}[htbp]
\scriptsize
\caption{RANSAC-Inspired Facade Detection Algorithm}\label{alg:ransac}
\begin{algorithmic}[1]
    \State \textbf{Input:} A set of $N$ \emph{aligned} orthographic images $\set{I}=\{I_{1}, \cdots, I_{N}\}$
    \State \textbf{Output:} Optimal consensus bounding box $b^*$ and Cropped facades $\set{F}=\{F_{1},\cdots, F_{N}\}$

    \Statex \textbf{Stage 1: Detect Lines and Create Proposal Set}
    \State Detect lines $\set{L}_k$ in each image $I_k \in \set{I}$ using the Line Segment Detector (LSD). Keep track of all detected lines $\{\set{L}_k\}_{k=1}^N$.
    \State Process all detected lines by filtering and merging them (both within each image and across images) to create a reliable line proposal set $\set{L}_r$. \Comment{This set represents line segments consistently identified across multiple views.}

    \Statex \textbf{Stage 2: RANSAC to Find Candidate Box per Image}
    \State $\set{B}_{candidates} \gets \text{empty list}$ \Comment{Store best box found for each image}
    \For{$k \gets 1$ to $N$} \Comment{Run RANSAC independently for each image $I_k$}
        \State Initialize best score $S_{best\_k} \gets -\infty$ and best box $b_k^* \gets \text{null}$ for image $I_k$.
        \State Let $NumIterations$ be the number of RANSAC trials. \Comment{Define RANSAC parameter}
        \For{$i \gets 1$ to $NumIterations$} \Comment{Inner RANSAC loop}
            \State Propose a candidate bounding box $b_{i,k}$ using two \emph{randomly selected} lines from the proposal set $\set{L}_r$.
            \State Calculate a score $S(b_{i,k})$ for the candidate box based on how well lines \emph{detected within the current image $I_k$} ($\set{L}_k$) fit within $b_{i,k}$. \Comment{Lines fully inside increase score, lines crossing decrease score.}
            \If{$S(b_{i,k}) > S_{best\_k}$}
                 \State Update $S_{best\_k} \gets S(b_{i,k})$ and $b_k^* \gets b_{i,k}$.
            \EndIf
        \EndFor
        \State $\set{B}_{candidates}.\text{append}(b_k^*)$ \Comment{Store the best box found for image $I_k$}
    \EndFor

    \Statex \textbf{Stage 3: Determine Final Consensus Bounding Box}
    \State Analyze the geometric properties (corner coordinates) of all candidate boxes stored in $\set{B}_{candidates}$.
    \State Determine the final consensus bounding box $b^*$ by clustering the candidate corners (using DBSCAN) and selecting the median coordinates of the dominant cluster.

    \Statex \textbf{Stage 4: Crop Facades}
    \State $\set{F} \gets \text{empty list}$ \Comment{Initialize list for cropped facade images}
    \For{$k \gets 1$ to $N$}
        \State $F_k \gets \text{CropImage}(I_k, b^*)$ \Comment{Crop image $I_k$ using the final consensus box $b^*$}
        \State $\set{F}.\text{add}(F_k)$
    \EndFor

    \State \textbf{return} $b^*$, $\set{F}$
\end{algorithmic}
\end{algorithm}

\begin{figure*}[h!]
\centering
\includegraphics[width=1\linewidth]{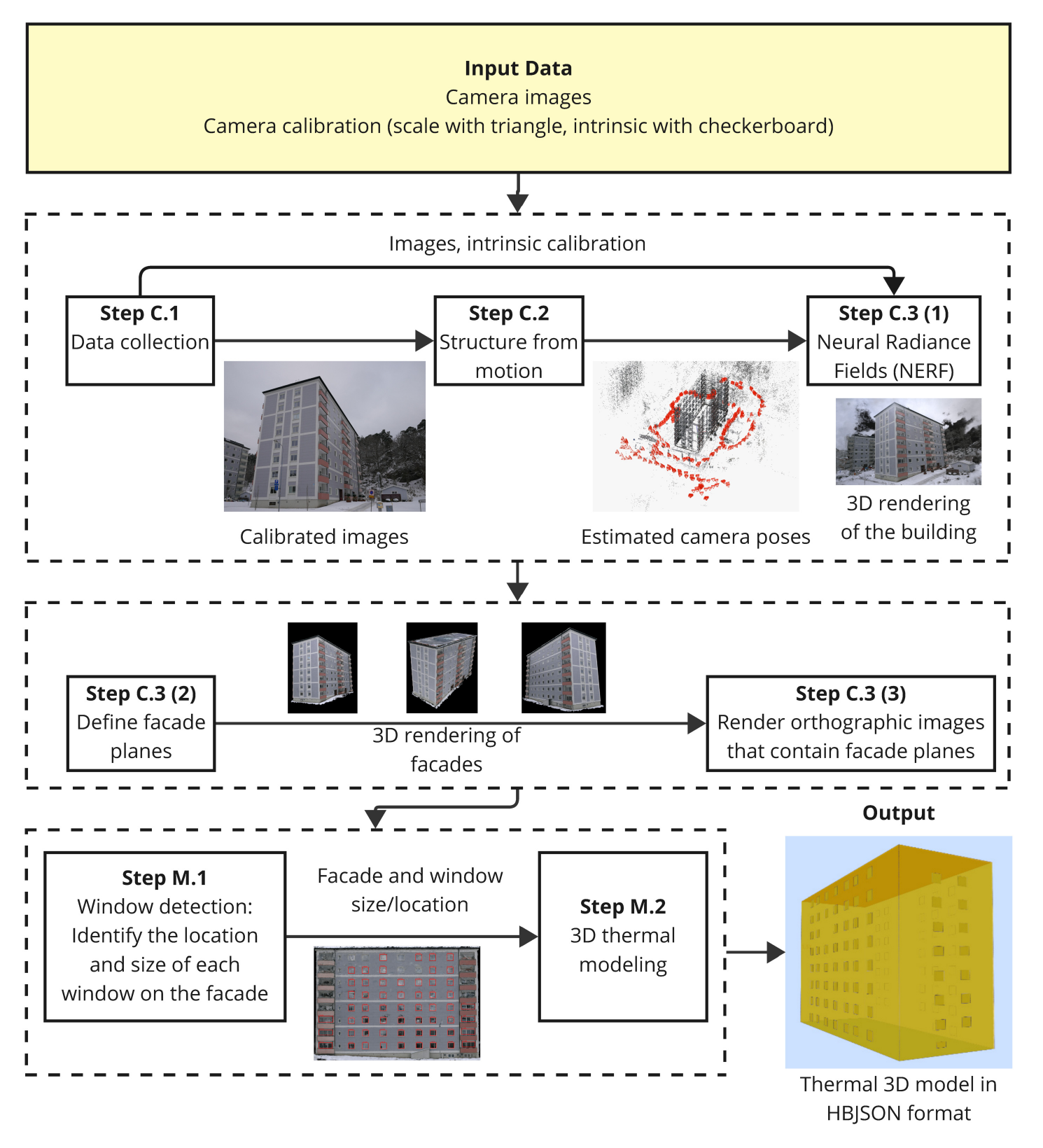}
\caption{\label{fig:workflow-nerf}  Workflow for the Camera2D pipeline.}
\end{figure*}

Facade detection can be achieved by deep learning techniques - we can opt for either Segment Anything \citep{kirillov2023segment} to segment any part of the image, deep neural networks trained for architectural features segmentation \citep{liu2020deepfacade} or object detection networks \citep{kong2020enhanced}. Given the complex scenarios encountered in large-scale datasets, such as obstructive trees or viewing angles that capture the side of a building, segmentation methods are not sufficient when precision is required not just at the pixel level but for comprehensively capturing entire objects, which is essential for subsequent 3D modeling. Bounding box detection, on the other hand, aligns with our needs. However, given the architectural diversity of buildings, a tailored dataset might be necessary for optimal results for each geographic locations. The geometry-based iterative RANSAC method proves to be sufficiently effective in managing such challenges.

\paragraph{Limitations}
The proposed path offers a systematic approach to achieve robust 3D semantic analysis for thermal modeling at scale.
However, large-scale data collection poses inevitable challenges. In areas where panoramic images are sparse, missing, or outdated, the accuracy of the 3D reconstruction may be compromised. Additionally, the process of aligning images and detecting facades can be hindered by environmental factors such as obstructive vegetation, poor lighting conditions, and suboptimal angles of capture. These factors can lead to inaccuracies in the plane definitions and alignments, which may affect the final 3D models and semantic analyses.
Moreover, our facade detection algorithm assumes that lines on the facade are either vertical or horizontal - its effectiveness can vary based on the architectural diversity of the buildings being analyzed.
Lastly, collecting and storing data at a large scale poses its unique challenges. To manage and make extensive datasets like GSV searchable, it is standard practice to parametrize the data and store only the parameters. For example, 3D objects (originally collected as LiDAR point clouds) in GSV are often parameterized and stored as flat planes. This method greatly enhances data management efficiency but introduces certain limitations to our modeling capabilities.
One significant artifact of this simplification is the occurrence of parallax errors. Parallax arises when objects are viewed from different angles, leading to apparent shifts in their positions relative to each other. In the context of facade modeling, using flat planes means that these positional shifts cannot be accurately captured, especially at larger viewing angles, where the effect is more significant. This simplification can distort the spatial relationships and dimensions in our models and affects the accuracy of the outputs.

Instead of depending on specific data provider such as GSV, alternative data collection methods can be considered. One approach is the use of mobile mapping systems (MMS), or unmanned aerial vehicles (UAVs), or drones equipped with location sensors, image cameras, and LiDAR. These equipment can capture high-resolution images from multiple angles and elevations as an alternative or complementary data collection method for the Streeview path.
Note that when collecting in-house data, post-processing and parameterization steps need to be carried out.

\subsection{Camera2D (C) for targeted inspection}
\label{sec:nerf_pipeline}

Despite the enhanced robustness and scalability of the StreetView path, challenges persist due to the broad approach to data collection.
For detailed examinations of a particular building of interest, the StreetView path might prove inadequate due to occlusion and missing facades.
To cover such use cases, we developed a second alternative path that targets the estimation of specific buildings to ensure a dense and complete data collection.
This methodology reconstructs 3D information via Neural Radiance Fields (NeRF), SfM, and real-world local reference points, aiming to mitigate common photogrammetric distortions for accurate digital twin generation.
By analyzing multiple photographs of a scene from different viewpoints, such techniques identify common keypoints and matches between images, which are then used to triangulate the 3D position of these points, forming a sparse 3D cloud.
Following this, it can perform dense reconstruction to get a more detailed 3D model.
Note that in addition to camera position, accurate camera orientation information can enhance the accuracy of the outcome, otherwise camera orientation can be estimated from the image data.
An overview can be found in Figure~\ref{fig:workflow-nerf}. There are three main steps (C.1-C.3) described as follows.

\begin{figure*}[ht!]
  \centering
  \includegraphics[width=0.9\linewidth]{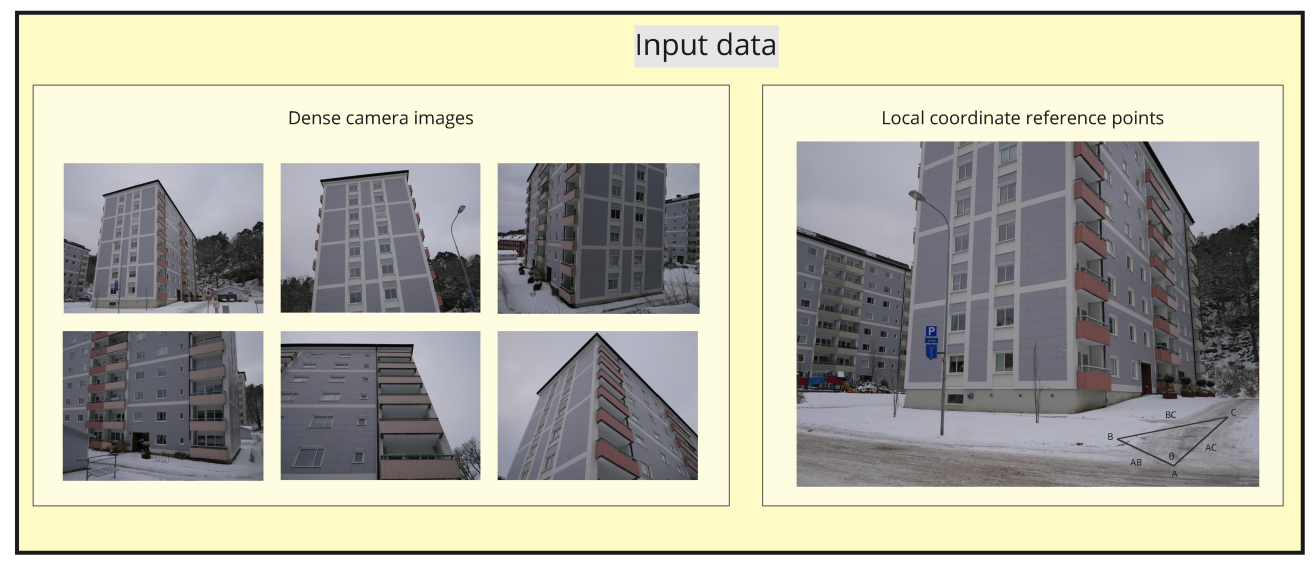}
  \caption{\label{fig:inputs-nerf} Input data for Camera2D.}
\end{figure*}

\paragraph{Step C.1: Data collection protocol}
Data collection for the Camera2D path involved capturing a dense set of photographs (typically 300 to 500 images) for each building, focusing on achieving sufficient facade coverage from multiple angles, aiming for 20\% overlap between sequential images, and incorporating loop closure where practical to enable robust 3D reconstruction via SfM and NeRF.  
The input data is illustrated in Figure~\ref{fig:inputs-nerf}.
Specifically, in our image acquisition protocol, we collect high-resolution still photographs due to their immunity to the rolling shutter effect and motion blur -- artifacts commonly associated with video capture \citep{liang2008analysis, ma2015handling}. These artifacts can significantly degrade the quality of data used in SfM processes, leading to less accurate 3D models.
In particular, we capture a dense array of photographs from multiple angles, which is critical for creating an overlapping dataset that enables a robust input with adequate redundancy for the SfM analysis. The redundancy not only enhances the precision of the resulting 3D model but also provides a safeguard against potential data loss or corruption.

Further, the photographs are taken in a way that they envelop the entire object, capturing every aspect and detail necessary for a complete reconstruction. The shooting angles and positions are guided by both the geometrical considerations of NeRF and the practical constraints of on-site conditions.

Moreover, as described in Section~\ref{sec:camera2d_3d}, images are taken from three local coordinate reference points (a real-world triangle on the ground) in proximity to the object.
The sides of the triangle are measured with an 8-meter measuring tape.
This triangle acts as a reference to anchor the scale for the entire sequence of the reconstruction process.
While an 8-meter tape is sufficient for our current requirements, alternative measurement tools, such as laser distance meters, could be used to potentially enhance the signal-to-noise ratio and measurement precision.

\paragraph{Step C.2: Camera pose estimation using COLMAP}

After image acquisition, we apply COLMAP \citep{schoenberger2016sfm} to reconstruct the 3D structure of the building from overlapping images and determining the intrinsic and extrinsic camera parameters.
COLMAP is a commonly used photogrammetry tool that automates 3D reconstruction from unordered image sets by integrating SfM and MVS techniques. It performs camera calibration, image matching, 3D model generation, and texture mapping. The process begins with detecting keypoints and extracting descriptors for each image, followed by matching features across images using appearance-based methods combined with geometric verification. Verified matches are used to construct a scene graph, where nodes represent images and edges represent shared keypoints. COLMAP then initiates an incremental reconstruction, starting from an image pair with sufficient matches and geometric diversity, progressively registering new images and triangulating additional points. Finally, a global bundle adjustment is performed to optimize camera poses and 3D point coordinates by minimizing the overall reprojection error. Through this pipeline, COLMAP produces detailed and accurate 3D models from large, unordered image collections.
Throughout this procedure, the intrinsic camera model parameters, which define the projection characteristics of the camera, are refined alongside the reconstruction.
In addition, during the SfM process, the camera poses corresponding to the triangle's corners are identified. The known lengths of the triangle’s sides are then applied to triangulate and determine the relative positions of these poses in metric units.
This step translates the relative scale derived from the SfM process into an absolute scale applicable to the real world, establishing a consistent and accurate scale across our SfM reconstruction.

\paragraph{Step C.3: NeRF modeling and orthographic projection}

To generate orthographic views from the inherently perspective NeRF model (Instant-NGP), we first define the target facade plane using {corner vertices (e.g., $\mathbf{v}_0, \mathbf{v}_1, \mathbf{v}_2$) of the relevant bounding box face} identified within the scaled 3D reconstruction (derived from Step C.2). These vertices define the plane's origin $\mathbf{v}_0$ and its basis vectors within the plane (e.g., $\mathbf{u} = \mathbf{v}_1-\mathbf{v}_0$, $\mathbf{v} = \mathbf{v}_2-\mathbf{v}_0$).
The orthographic image is then generated by densely sampling the NeRF volume. For each pixel $(x, y)$ in the target orthographic image (of pixel dimensions $W \times H$), the corresponding 3D point $\mathbf{p}(x,y)$ on the facade plane is calculated:
\begin{equation} \label{eq:OrthoPoint}
\mathbf{p}(x,y) = \mathbf{v}_0 + \frac{x}{W} \mathbf{u} + \frac{y}{H} \mathbf{v} 
\end{equation}

\begin{algorithm}[htbp] 
\scriptsize 
\caption{NeRF-based Orthographic Projection}\label{alg:nerf-ortho}
\begin{algorithmic}[1] 
	\State \textbf{Input:} Trained NeRF model $\set{M}_{NeRF}$, Facade corners $\set{P}_{corners}$, Pixel size $s_{pixel}$, Samples per pixel $N_{spp}$
	\State \textbf{Output:} Orthographic image $\mat{I}_{ortho}$

	\Statex \textbf{Stage 1: Define Plane Geometry \& Coordinate System}
	\State $\vect{p}_{origin} \gets \set{P}_{corners}[0]$
	\State $\vect{u}, \vect{v}, \vect{w} \gets \text{CalculatePlaneBasis}(\set{P}_{corners})$ \Comment{$\vect{w}$ is normal}
	\State $\mat{R}_{ortho} \gets \text{RotationMatrixFromBasis}(\vect{u}, \vect{v}, \vect{w})$

	\Statex \textbf{Stage 2: Determine Output Resolution}
	\State $W_{facade} \gets \text{Distance}(\set{P}_{corners}[1], \vect{p}_{origin})$
	\State $H_{facade} \gets \text{Distance}(\set{P}_{corners}[2], \vect{p}_{origin})$
	\State $W_{out} \gets \text{Round}(W_{facade} / s_{pixel})$
	\State $H_{out} \gets \text{Round}(H_{facade} / s_{pixel})$
	\State $\mat{I}_{ortho} \gets \text{CreateImageBuffer}(W_{out}, H_{out})$

	\Statex \textbf{Stage 3: Generate Orthographic Image Pixels}
	\For{$y \gets 0$ to $H_{out} - 1$}
    	\For{$x \gets 0$ to $W_{out} - 1$}
        	\State $\vect{p}_{world} \gets \vect{p}_{origin} + (y \cdot s_{pixel} \cdot \vect{v}) + (x \cdot s_{pixel} \cdot \vect{u})$
        	\State $\mat{T}_{c2w} \gets \text{SetupOrthoCameraAtPoint}(\vect{p}_{world}, \mat{R}_{ortho})$
        	\State $\text{SetNeRFCamera}(\set{M}_{NeRF}, \mat{T}_{c2w}, \text{FOV} \approx 0)$
        	\State \Comment{Core step: Synthesize pixel color from 3D model}
        	\State $C_{pixel} \gets \text{RenderFromNeRF}(\set{M}_{NeRF}, N_{spp})$
        	\State $\mat{I}_{ortho}[y, x] \gets C_{pixel}$
    	\EndFor
	\EndFor

	\State \textbf{return} $\mat{I}_{ortho}$
\end{algorithmic}
\end{algorithm}

The NeRF model is then queried to render the color and depth information along a \textbf{single ray originating at $\mathbf{p}(x,y)$ and directed perpendicularly outwards} from the facade plane. Assembling these rendered samples pixel-by-pixel simulates parallel projection, creating the orthographic image. Key configuration parameters included setting the axis-aligned bounding box scale (to 4, in our case). Training proceeded until convergence based on visual inspection of rendering quality and stabilization of the training loss.
The algorithm is described in Algorithm~\ref{alg:nerf-ortho}.
This method directly renders the full extent of the defined facade plane at the desired real-world scale determined in Step C.2. It renders the scene including any foreground occlusions reconstructed by NeRF but avoids occlusion issues associated with distant virtual cameras.

\paragraph{Limitations}
The ground-based camera angle limits the visibility of roof structures. Further, achieving a successful SfM reconstruction requires a significant number of images. Since images are typically taken from ground level, this can introduce perspective distortion in higher parts of buildings, making the rendering noisier in orthographic projections.
An alternative might involve using drones for capturing images, which can directly provide camera poses and potentially 3D information through LiDAR, though at a higher cost.

Furthermore, while we render NeRF images as orthographic projections to standardize the image representation, we acknowledge that their underlying pixel distributions can still differ substantially from real-world images. This difference may pose challenges for object detection and classification, as models trained only on real-world data may not generalize effectively to NeRF-rendered inputs. We did not apply any domain adaptation or special training strategies to mitigate this gap in our paper. Addressing this limitation remains an important direction for future work.

\subsection{Merged steps}
\label{sec:merge}
The aforementioned steps for both alternative paths, StreetView and Camera2D, produce true-to-scale orthographic facades. To complete the 3D thermal modeling, two remaining merged steps are described as follows.

\paragraph{Step M.1: Semantic facade parsing} Identify the location and size of each window on the facade.
Once the facade is identified, a pretrained ResNet-50 RetinaNet, trained on the LSAA dataset \citep{9145640}, is used for window detection.
More specifically, a ResNet-50 RetinaNet model as the base model, initialized with weights pre-trained on the COCO 2017 dataset, was fine-tuned for 64,000 iterations specifically for this task using the LSAA dataset (Zhu et al., 2020) of architectural facade elements. The training is conducted using an SGD optimizer with a learning rate of 0.0005 and data augmentation incorporating color jitter and random resizing/scale jittering (based on 1440px max resize, 800-1600px scale range). For inference, orthographic images were resized (maintaining aspect ratio, shortest side to 1024px, longest side max 1333px) and normalized using standard ImageNet statistics.
If multiple images are available for the same facade, detections from individual views are fused using a specialized ensemble method (Algorithm \ref{alg:detection_fusion}) for enhanced robustness. Initially, only detections exceeding a confidence score of 0.2 are considered. Viewpoints are not explicitly weighted. This involves grouping these filtered detections from different views based on high spatial overlap (IoU $>$ 0.3) to identify potential matches corresponding to the same physical window. To ensure reliability and consistency, providing robustness against partial occlusions and spurious detections, only groups representing windows detected across at least two different source images (N $\geq$ 2) are considered valid. For each valid group, geometric information and confidence scores are merged to generate a single, representative detection. This merged detection must also meet a final score threshold $\tau_{score2}$, set to 0.4, to be accepted. This algorithm is described in Algorithm~\ref{alg:detection_fusion}.

\begin{algorithm}[htbp]
\scriptsize
\caption{StreetView Multi-View Detection Fusion}\label{alg:detection_fusion}
\begin{algorithmic}[1]
    \State \textbf{Input:} Dictionary $\set{D}_{raw}$ mapping source $k=(pano\_id, plane\_idx)$ to lists of raw detections $d = \{\text{bbox}, \text{score}, \text{category\_id}\}$, Confidence threshold $\tau_{conf}$, IoU threshold $\tau_{iou}$, Minimum detections per cluster $N_{min}$, Merged score threshold $\tau_{score2}$
    \State \textbf{Output:} List $\set{D}_{final}$ of merged and validated detections for the facade.

    \Statex \textbf{Stage 1: Filter Raw Detections by Confidence}
    \State $\set{L}_{det} \gets \text{empty list}$ \Comment{Store filtered detections meeting initial confidence}
    \For{$k \in \text{Keys}(\set{D}_{raw})$} \Comment{Iterate through each source view}
        \For{$d \in \set{D}_{raw}[k]$} \Comment{Iterate through detections in the view}
            \If{$d.\text{score} \ge \tau_{conf}$}
                \State $\set{L}_{det}.\text{append}(d)$ \Comment{Keep detection if score is high enough}
            \EndIf
        \EndFor
    \EndFor
    \If{$\text{Length}(\set{L}_{det}) == 0$} \textbf{return} empty list \EndIf \Comment{Exit if no detections pass initial filter}

    \Statex \textbf{Stage 2: Group by Category and Cluster Spatially}
    \State $\set{D}_{final} \gets \text{empty list}$ \Comment{Initialize final output list}
    \For{$cat\_id \in \text{UniqueCategories}(\set{L}_{det})$} \Comment{Process one object category at a time}
        \State $\set{D}_{cat} \gets \text{DetectionsInCategory}(\set{L}_{det}, cat\_id)$ \Comment{Get all filtered detections of this category}
        \If{$\text{Length}(\set{D}_{cat}) < 2$} \textbf{continue} \EndIf \Comment{Need at least two detections to potentially form a cluster}

        \State \textit{// Cluster detections based on bounding box overlap (IoU)}
        \State $\textit{Clusters}_{idx} \gets \text{ClusterByHighIoU}(\set{D}_{cat}, \tau_{iou})$ \Comment{Groups indices of detections that likely represent the same object instance}

        \Statex \textbf{Stage 3: Validate and Merge Spatial Clusters}
        \For{$idx_{cluster} \in \textit{Clusters}_{idx}$} \Comment{Process each spatial cluster}
            \State \textit{// --- Validation 1: Check Minimum Detections in Cluster ---}
            \If{$\text{Length}(idx_{cluster}) < N_{min}$} \Comment{Check if cluster has enough total detections}
                \State \textbf{continue} \Comment{\textbf{Reject cluster}: Too few detections grouped together}
            \EndIf

            \State \textit{// --- Merge Cluster ---}
            \State $\set{M}_{det} \gets [\set{D}_{cat}[i] \text{ for } i \in idx_{cluster}]$ \Comment{Get the actual detection objects in this cluster}
            \State $d_{final} \gets \text{MergeDetectionsSpecific}(\set{M}_{det})$ \Comment{Compute representative bbox (median coords) and score (mean of sqrt scores)}
            \State $d_{final}.\text{category\_id} \gets cat\_id$ \Comment{Assign the category ID}

            \State \textit{// --- Validation 2: Check Merged Score Threshold ---}
            \If{$d_{final}.\text{score} < \tau_{score2}$} \Comment{Check if the merged detection is confident enough}
                \State \textbf{continue} \Comment{\textbf{Reject cluster}: Merged score too low}
            \EndIf

            \State \textit{// --- Store Validated and Merged Detection ---}
            \State $\set{D}_{final}.\text{append}(d_{final})$ \Comment{Keep the final detection for this cluster}
        \EndFor
    \EndFor

    \State \textbf{return} $\set{D}_{final}$ \Comment{Return the list of high-confidence, merged detections}
\end{algorithmic}
\end{algorithm}

 This method enhances the reliability of the detection process by ensuring consistency across different images.
 Once windows are detected and their positions within the facade are determined, the dimensions in these elements in the image can be translated into real-world measurements by leveraging the respective scale information (i.e. plane definition for StreetView and local coordinate reference points for Camera2D; cf.~Section~\ref{sec:camera2d_3d}).
These outcomes collectively provide users with the essential data needed to reconstruct the facade in 3D.

\paragraph{Step M.2: 3D thermal modeling} Once the windows are detected, their locations and scales can be used to render them in 3D. When combined with available footprint information, a 3D model of the complete building can be reconstructed using HoneybeeJSON \footnote{\url{https://github.com/ladybug-tools/honeybee-schema}}, a standardized JSON schema to encode geometric information about the building's envelope, including the coordinates of the facades and the location of windows. Once the geometric information is encoded into the HoneybeeJSON schema, thermal properties of the building, such as material properties and the u-value of the windows. The input of these properties could be automated using databases such as TABULA \citep{Loga2016} or Energy Performance Certificates, which is beyond the scope of this paper. The Honeybee model can be simulated in EnergyPlus using the built-in translation tools to provide the energy demand of the building.

\begin{figure}[h!]
\centering
\includegraphics[width=1\linewidth]{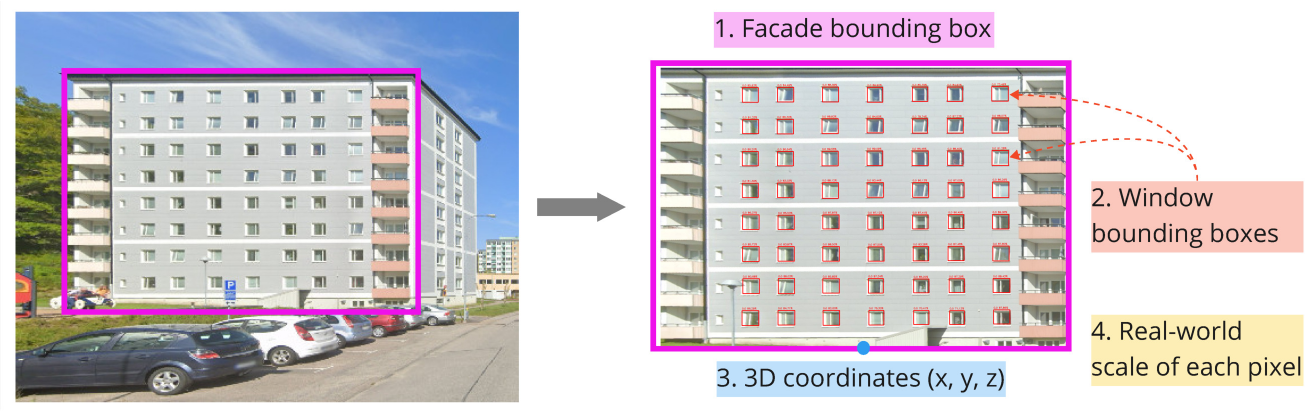}
\caption{\label{fig:outcome} Final unified outputs of the two alternative paths, Camera2D and StreetView, in the SI3FP pipeline.}
\end{figure}

\paragraph{The complete SI3FP pipeline}
The complete pipeline can be found in Algorithm~\ref{alg:si3fp_pipeline_final}.
\begin{algorithm}[htbp] 
\scriptsize 
\caption{SI3FP Pipeline}\label{alg:si3fp_pipeline_final} 
\begin{algorithmic}[1]
    \State \textbf{Input:} Data Source Type `DataSourceType` (`Camera2D` or `StreetView`), `InputData`, `BuildingFootprint`, Target pixel size $s_{pixel}$, Detection thresholds ($\tau_{conf}, \tau_{iou}, N_{min}, \tau_{score2}$)
    \State \textbf{Output:} Thermal 3D model `ThermalModel` (HBJSON format)

    \If{`DataSourceType` == `Camera2D`}
        \Statex \textbf{--- Camera2D Path ---}
        \State \textit{// C1: Data Collection}
        \State $\set{I}_{dense} \gets \text{InputData}$ \Comment{Assume dense image data collected}
        \State \textit{// C2: Structure-from-Motion}
        \State $\set{T}_{cam}, \set{P}_{corners} \gets \text{EstimateScaledPosesAndFacadeCorners}(\set{I}_{dense})$ \Comment{Using COLMAP}
        \State \textit{// C3a: NeRF Training}
        \State $\set{M}_{NeRF} \gets \text{TrainNeuralRadianceField}(\set{I}_{dense}, \set{T}_{cam})$ \Comment{Using Instant-NGP}
        \State \textit{// C3b: Orthographic Projection}
        \State $\mat{I}_{ortho} \gets \text{GenerateOrthoNeRF}(\set{M}_{NeRF}, \set{P}_{corners}, s_{pixel})$ \Comment{Using Alg \ref{alg:nerf-ortho}}
        \State $G_{facade} \gets \text{GetFacadeGeometryFromCorners}(\set{P}_{corners})$
        \State $\set{F} \gets \{\mat{I}_{ortho}\}$ \Comment{Define Facade image set}

    \ElsIf{`DataSourceType` == `StreetView`}
        \Statex \textbf{--- StreetView Path ---}
        \State \textit{// S1: Data Collection}
        \State $\set{P}_{anos} \gets \text{InputData}$ \Comment{Assume panorama data collected}
        \State \textit{// S2: Plane Clustering}
        \State $\set{C}_{geom} \gets \text{ClusterStreetViewPlanes}(\set{P}_{anos})$ \Comment{Using Alg \ref{alg:plane_clustering}}
        \State \textit{// Generate Per-Pano Ortho Images}
        \State $\set{I}_{ortho\_segments} \gets \text{GeneratePerPanoOrthographics}(\set{C}_{geom}, \set{P}_{anos}, s_{pixel})$ \Comment{Alg \ref{alg:per_pano_ortho_math}; uses Alg \ref{alg:streetview-ortho} logic}
        \State \textit{// S3: Image Alignment}
        \State $\set{I}_{aligned} \gets \text{AlignOrthographicImages}(\set{I}_{ortho\_segments})$ \Comment{Using SIFT/Registration}
        \State \textit{// S4: Facade Detection}
        \State $b^*, \set{F} \gets \text{DetectFacadeBoundsAndCrop}(\set{I}_{aligned})$ \Comment{Using Alg \ref{alg:ransac} logic}
        \State $G_{facade} \gets \text{GetFacadeGeometryFromBbox}(b^*)$
    \EndIf

    \Statex \textbf{--- Merged Steps ---}
    \State \textit{// M1: Semantic Facade Parsing}
    \State $\set{D}_{raw} \gets \text{DetectWindowsInitial}(\set{F})$ \Comment{Apply detector (RetinaNet) to image(s) in $\set{F}$}
    \If{`DataSourceType` == `StreetView`}
        \State $\set{D}_{final} \gets \text{FuseMultiViewDetections}(\set{D}_{raw}, \tau_{conf}, \tau_{iou}, N_{min}, \tau_{score2})$ \Comment{Using Alg \ref{alg:detection_fusion}}
    \Else
        \State $\set{D}_{final} \gets \text{FilterSingleViewDetections}(\set{D}_{raw}, \tau_{conf})$ \Comment{Camera2D case}
    \EndIf
    \State $G_{windows} \gets \text{CalculateWindowGeometries3D}(\set{D}_{final}, G_{facade}, s_{pixel})$

    \State \textit{// M2: 3D Thermal Model Generation}
    \State $ThermalModel \gets \text{AssembleHBJSONModel}(G_{facade}, G_{windows}, BuildingFootprint)$

    \State \textbf{return} $ThermalModel$
\end{algorithmic}
\end{algorithm}
\paragraph{Final output}
The final output is a thermal 3D model in HBJSON format.
More specifically, for each facade, the outcome consists of four key components illustrated in Figure~\ref{fig:outcome}.
\begin{itemize}
\item[1.] Facade bounding box: Each path generates an orthographic image of the facade, with a bounding box describing its position and dimension within the image.
\item[2.] Window bounding boxes: Surrounding all windows within the facade, bounding boxes are detected to detail their positions and dimensions.
\item[3.] 3D coordinates of the facade center: The central point of the facade's bounding box is pinpointed in 3D space, specified by its latitude, longitude, and altitude.
\item[4.] Real-world scale of each pixel: The real-world scale of each pixel within the image is calculated from the scale to enable measurements from the image to actual dimensions.
\end{itemize}

\section{Experiments and Results}
This section outlines the experimental setup for the evaluation of the alternative paths in the pipeline SI3FP.

\subsection{Experimental Setup}
\label{sec:experimental_setup}
\begin{figure}[h!]
\centering
\includegraphics[width=1\linewidth]{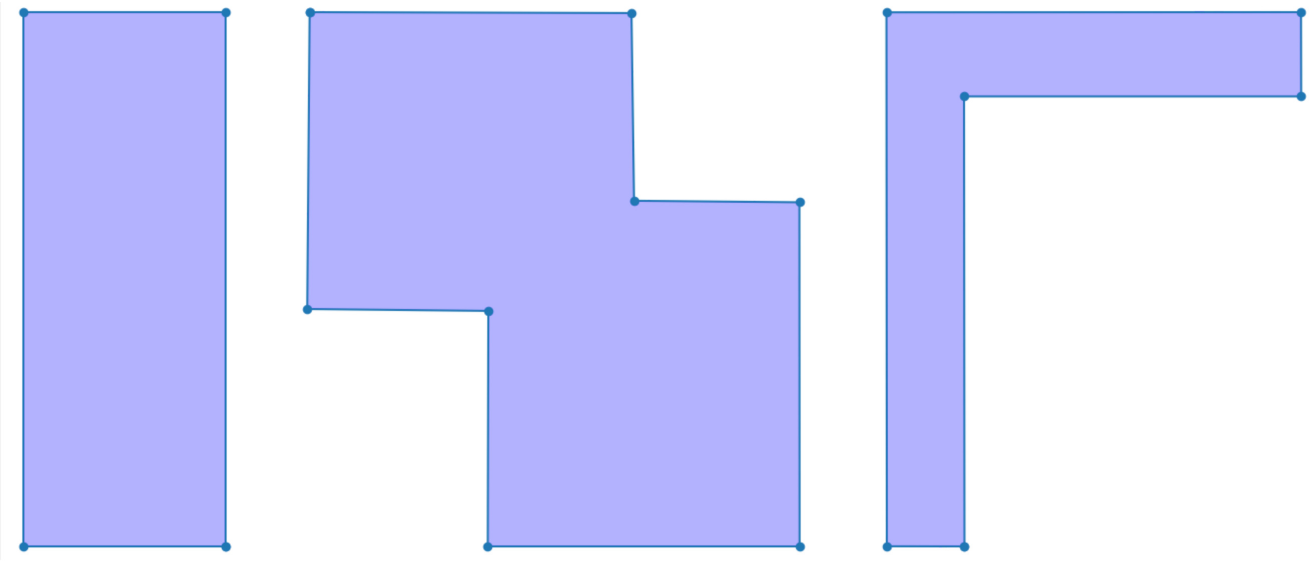}
\caption{\label{fig:footprint} Footprints of the buildings in our case study.}
\end{figure}
\paragraph{Data collection}
To test and validate our pipeline with real-world data, we collected ground truth from three buildings in Sweden. We use three multi-family residential buildings from 1961 to 1975, because buildings from this period urgently need energy renovation \citep{MANGOLD201641}. This era is known as the Million Homes Program -- a national initiative to rapidly build one million dwellings to deal with the housing shortage \citep{HALL01012005}.  Multi-family buildings represent 54\% of the dwellings in the Swedish residential building stock \citep{SAVVIDOU2020111679}.   
These typologies are common not only in Sweden but all over Europe and represent 36\% of the European residential building stock by floor area \footnote{\url{https://www.bpie.eu/wp-content/uploads/2015/10/HR_EU_B_under_microscope_study.pdf}}. 
The footprints can be found in Figure~\ref{fig:footprint}.
For the StreetView path, approximately 15-40 relevant panoramic views per building facade cluster, obtained via the GSV API and filtered as described in Section\ref{sec:streeview_pipeline}.
For the Camera2D path, the dense sets of ~4 MP photographs captured for each building, with the detailed capture protocol described in Step C.1, Section\ref{sec:nerf_pipeline}. The information and the total number of images collected for each building is detailed in Table~\ref{tab:building_summary}. 
\begin{table}[ht]
\centering
\caption{Dataset overview for each building.}\label{tab:building_summary}
\resizebox{\textwidth}{!}{
\begin{tabular}{lccc|cc|ccc}
\toprule
\multirow{2}{*}{Building} & \multicolumn{3}{c|}{Building Information} & \multicolumn{2}{c|}{Camera2D} & \multicolumn{3}{c}{StreetView} \\
\cmidrule(lr){2-4} \cmidrule(lr){5-6} \cmidrule(lr){7-9}
 & Height & Longest Side & Facades & Images & Resolution (px) & Panoramas & Resolution (px) & Missing Facades \\
\midrule
B1 & 23.2 & 14.9 & 8 & 322 & $2296\times1724$ & 47 & $16384\times8192$ & 3 (37.5\%) \\
B2 & 22.7 & 35.2 & 4 & 439 & $2250\times1680$ & 22 & $16384\times8192$ & 1 (25.0\%)\\
B3 & 11.0 & 71.6 & 6 & 274 & $2288\times1708$ & 58 & $16384\times8192$ & 2 (33.3\%)\\
\bottomrule
\end{tabular}
}
\end{table}
\paragraph{Evaluation}
In this paper, we focus on using the pipeline SI3FP for thermal 3D modeling, and hence we primary discuss around four evaluation aspects relevant to our application.
\begin{itemize}
\item Window detection:
First of all, we are interested in evaluating the window detection rate.
This is a critical intermediate step in facade parsing. A window prediction was considered a TP if its Intersection over Union (IoU) with the corresponding ground truth bounding box exceeded a threshold of 0.5, following the standard object detection practice \citep{Everingham10}. To evaluate the performance of window detection, we use the F1-Score, which is the harmonic mean of Precision ($P = TP/(TP+FP)$) and Recall ($R = TP/(TP+FN)$)\footnote{TP:True Positive; FP: False Positive; TN: True Negative; FN: False Negative.}. Note: `StreetView (Total)' considers all ground truth windows, while `StreetView' without `Total' only considers facades where the method produced predictions. 
\item Area of the windows and facades:
The total area covered by windows is an important parameter in thermal modeling, as it directly influences the building's heat gain and loss.
Accurate measurement of window areas enables calculations of thermal load and energy requirements. The area is evaluated by the {Mean Absolute Relative Area Error} (the average magnitude of the relative area error over matched window pairs ($A(w)$ is window area):
    $$ \text{Mean Abs Rel Area Err} = \frac{1}{|M|} \sum_{(w_p, w_{gt}) \in M} \left| \frac{A(w_p) - A(w_{gt})}{A(w_{gt})} \right| $$
\item Location of the windows:
Spatially accurate window location helps in assessing natural light distribution within interiors, which affects both energy consumption for lighting and heating.
The estimated location is evaluated by the {Mean Absolute Position Error (m):} (the average Euclidean distance between centers of matched window pairs, in meters):
    $$ \text{Mean Abs Pos Err (m)} = \frac{1}{|M|} \sum_{(w_p, w_{gt}) \in M} D(C(w_p), C(w_{gt})) $$
    
Further, the location and area are jointly evaluated by the \textbf{Mean  (IoU)}, which is defined as the average IoU over correctly matched window pairs $(w_p, w_{gt}) \in M$.
    $$ \text{Mean IoU} = \frac{1}{|M|} \sum_{(w_p, w_{gt}) \in M} IoU(w_p, w_{gt}) $$
\item WWR (window-to-wall ratio):
 Based on an expert workshop with three representatives of large residential building portfolio owners in West Sweden, we established that a 5\% error in WWR is accurate enough in the early planning phase. \cite{LU2023113275} confirmed this assumption and showed that experts have higher deviations in their estimations in this phase. 
The impact a 5\% error in WWR has on the simulated energy demand depends on many factors. Assuming a WWR of 0.25, an increase of 5\% would correspond to an increase of 0.75 \% in heating demand according to a study in Sweden \citep{etde_635147}, which is acceptable in this phase according to the expert workshop. 

More specifically, we evaluate the WWR in the following manner:
\begin{itemize}
    \item {WWR Error without missing facades (e.g. Camera2D and StreetView Standard):} The mean of per-facade errors, $e(j) = \hat{WWR}(j) - WWR_{gt}(j)$, averaged over the set of facades $F_{pred}$ where the method produced predictions. $\hat{WWR}(j)$ is the actual predicted WWR for facade $j$.

    \item {WWR Error with missing facade (e.g. StreetView Total):} The mean of per-facade errors, $e(j) = \hat{WWR}(j) - WWR_{gt}(j)$, averaged over \textit{all} ground truth facades $F_{gt}$. If a facade $j$ is missing, its estimated $\hat{WWR}(j)$ is treated as $0.0$.

    \item {Imputed WWR Error with missing facades (e.g. StreetView Total imputed):} The mean of per-facade errors, $e(j) = \hat{WWR}_{imp}(j) - WWR_{gt}(j)$, averaged over \textit{all} ground truth facades $F_{gt}$. If facade $j$ lacks a StreetView prediction, its estimated WWR $\hat{WWR}_{imp}(j)$ is imputed using the average WWR calculated from facades that \textit{did} have predictions ($\overline{WWR}_{pred}$).
\end{itemize}
\end{itemize}
Components like COLMAP and NeRF are applied or fitted to the data from these three buildings directly. The detection algorithm was fine-tuned on a separate, custom dataset of architectural elements (starting from COCO pre-trained weights). The three case study buildings were held out from this fine-tuning process, thus serving as an independent test set to evaluate the detector's generalization performance within the overall pipeline framework.

\paragraph{LiDAR scan as the ground truth}
We used the Topcon GLS-2000 scanner as the reference sensor to create the ground truth. The scanner offers a 360-degree scanning range. With a distance accuracy of 3.5 mm up to 150 meters and a horizontal and vertical angle accuracy of 6 arc-seconds, it produces dense point clouds, reaching up to 120,000 points per second. This collected data is stored as .pts file, a simple text file and loaded into the software Rhinoceros 3D as .xyz file. The geometry was manually modeled over the point cloud. In case of unclear areas, e.g. sparce points due to obstructions, architectural drawings and GSV images were referenced. Then the surface of the geometry was extracted and labeled with descriptions such as exterior wall, window, roof, etc., by using Grasshopper and the plugin Honeybee. The final model was exported as a JSON file and checked for correct interpretation in OpenStudio.

\paragraph{Computational infrastructure}
All data processing, model training (NeRF), and inference tasks reported in this paper were conducted on a workstation equipped with an AMD Ryzen 7 5800X3D CPU, 64GB of RAM, and an NVIDIA RTX 3090 GPU with 24GB of VRAM.

\subsection{Results}
The overall results are presented in Table~\ref{tab:window_detection}, evaluated across three buildings using Camera2D and StreetView data sources. We discuss the evaluation metrics below.

\begin{table}
\caption{Evaluation results of SI3FP on three buildings.}
\label{tab:window_detection}
\resizebox{\textwidth}{!}{
\begin{tabular}{llrrrrrrr}
\toprule
\thead{Building} & \thead{Method} & \thead{F1-Score} & \thead{F1-Score \\ (No Balc)} & \thead{Mean \\ IoU} & \thead{Mean Abs Rel \\ Area Err (\%)} & \thead{Mean Abs \\ Pos Err (m)} & \thead{WWR \\ Error} & \thead{WWR Error \\ (Imputed)} \\
\midrule
B1 & Camera2D & 0.75 & 0.90 & 0.80 & 17.1\% & 0.065 & -0.080 & - \\
B1 & StreetView & 0.75 & 0.97 & 0.70 & 15.2\% & 0.189 & -0.090 & - \\
B1 & StreetView (Total) & 0.71 & 0.65 & - & - & - & -0.095 & -0.070 \\
B2 & Camera2D & 0.83 & 0.87 & 0.78 & 21.9\% & 0.077 & -0.009 & - \\
B2 & StreetView & 0.80 & 0.81 & 0.73 & 23.2\% & 0.105 & -0.052 & - \\
B2 & StreetView (Total) & 0.61 & 0.63 & - & - & - & -0.092 & -0.042 \\
B3 & Camera2D & 0.72 & 0.84 & 0.77 & 15.2\% & 0.129 & -0.048 & - \\
B3 & StreetView & 0.28 & 0.46 & 0.69 & 22.3\% & 0.216 & -0.040 & - \\
B3 & StreetView (Total) & 0.19 & 0.04 & - & - & - & -0.099 & -0.048 \\
Total & Camera2D & 0.76 & 0.86 & 0.78 & 17.8\% & 0.095 & -0.038 & - \\
Total & StreetView & 0.60 & 0.82 & 0.71 & 19.2\% & 0.165 & -0.058 & - \\
Total & StreetView (Total) & 0.47 & 0.42 & - & - & - & -0.095 & -0.050 \\
\bottomrule
\end{tabular}
}
\end{table}

\paragraph{Window detection}

\begin{figure}[ht!]
\centering
\includegraphics[width=0.8\linewidth]{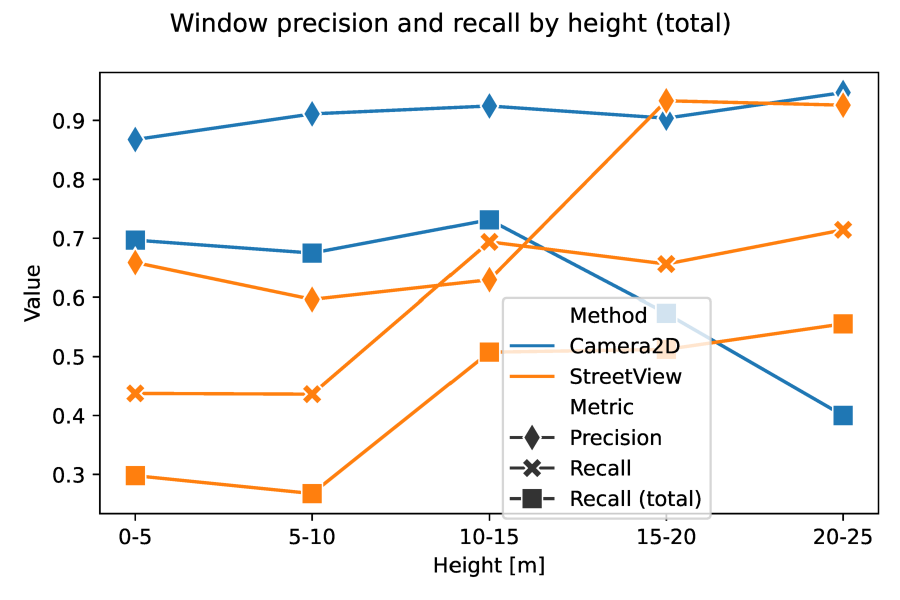}
\caption{\label{fig:Window_precision_and_recall_by_height} Precision and recall of window detection versus the height of the building.}
\end{figure}
The detection result is summarized by the F1-Score in Table~\ref{tab:window_detection}. 
 Overall, Camera2D consistently achieves better performance than StreetView across almost all evaluation metrics. This difference is mainly due to the higher control over data acquisition in Camera2D: images are captured specifically for the task with careful coverage and appropriate resolution. In contrast, StreetView imagery is captured opportunistically, often from a distance or obstructed viewpoints, leading to missing facades and less optimal angles.

 A clear trend emerges when comparing F1-Scores with and without balconies. Across all buildings, excluding facades with balconies significantly improves detection performance, particularly for StreetView imagery. Balconies introduce heavy occlusions, causing windows to be partially or entirely hidden. This makes reliable window detection extremely challenging, especially for automated methods trained on unobstructed data assumptions. 

 Building geometry influences performance for StreetView. B3, with its long facades, shows the worst results. Long facades require capturing windows across wide angles and distances in StreetView, which increases distortion and reduces the effective resolution for distant windows. 

As noted in Section~\ref{sec:nerf_pipeline}, the NeRF algorithm used in Camera2D requires high quality images to produce sharp renders. 
Cameras positioned at ground level and angled upwards often introduce significant perspective distortion.
This results in blurring in both the reconstructed 3D renderings and the orthographic images, particularly impacting the detection of windows at higher elevations (decrease in recall) especially when the windows are small.
\begin{figure}[h!]
\centering
\includegraphics[width=0.8\linewidth]{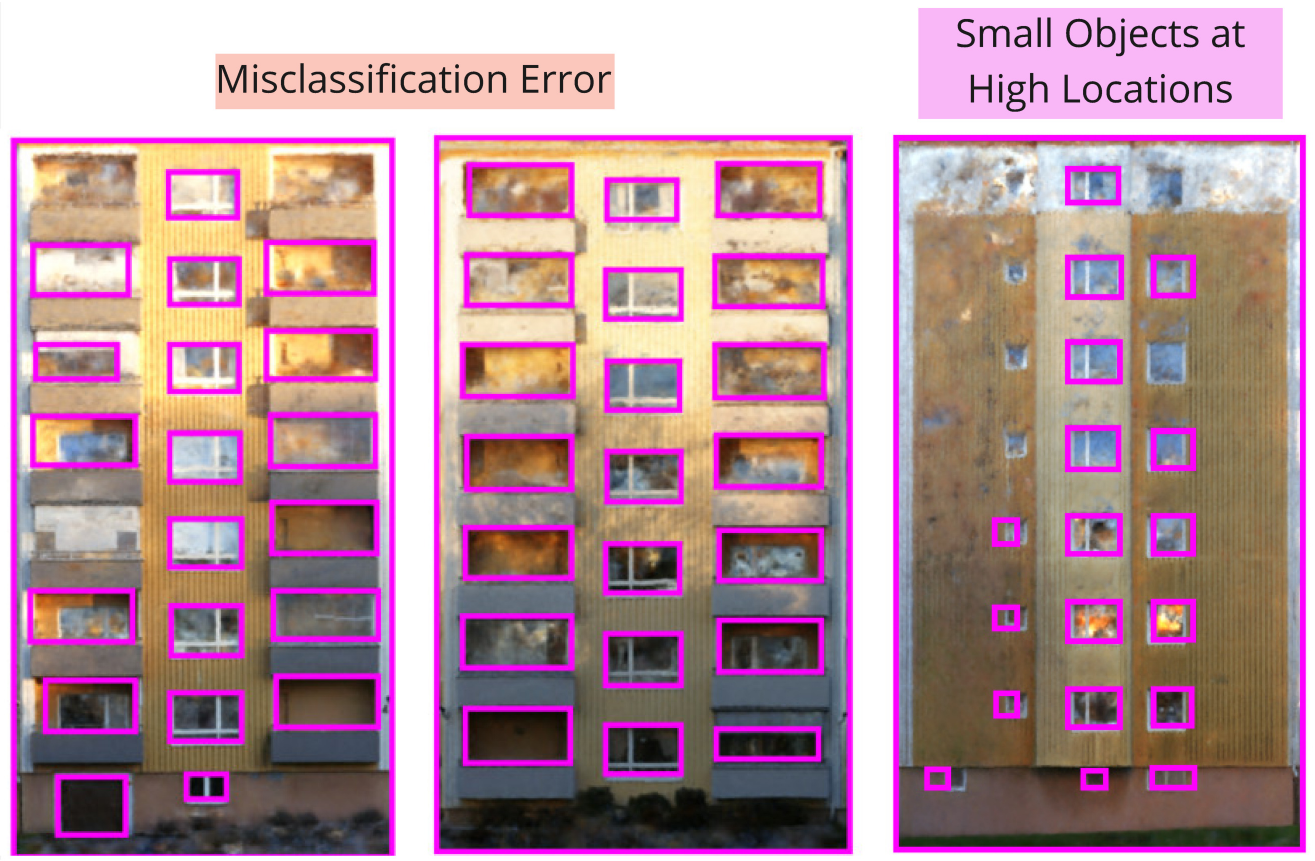}
\caption{\label{fig:nerf_example} Two types of errors in Camera2D.}
\end{figure}

One potential solution to this source of error is to employ drones equipped with video cameras, which can capture imagery from elevated and varied angles, reducing perspective issues.

Figure~\ref{fig:Window_precision_and_recall_by_height} illustrates how precision and recall vary with the height of the window position. Note that StreetView experiences an increase in both recall and precision at higher heights, because the lower parts of facades are more frequently obscured by obstacles such as vegetation and vehicles.

Another type of error is classification errors, where architectural features such as balconies with window-like structures are mistakenly identified as windows.
To address this, we can fine-tune our deep learning models on more specialized datasets.
While this approach demands additional resources -- being both time-consuming and potentially costly -- it could lead to improved accuracy tailored to specific use cases where such distinctions are critical.

Examples for qualitative visual inspection of these errors can be found in Figure~\ref{fig:nerf_example}.

\begin{figure}[ht!]
\centering
\includegraphics[width=0.8\linewidth]{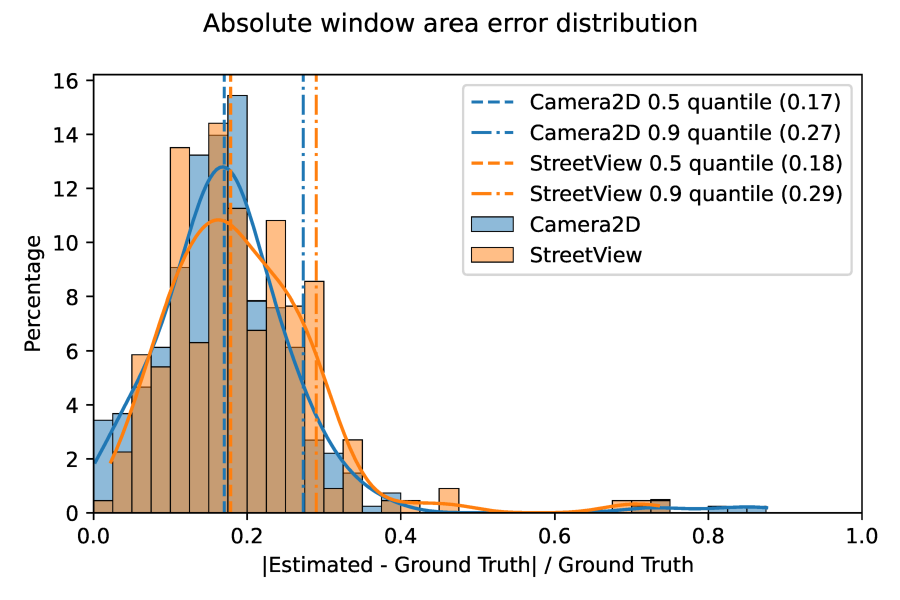}
\caption{\label{fig:Absolute_window_area_error_distribution} Distribution of absolute errors in window area estimations.}
\end{figure}
\begin{figure}[ht!]
\centering
\includegraphics[width=0.8\linewidth]{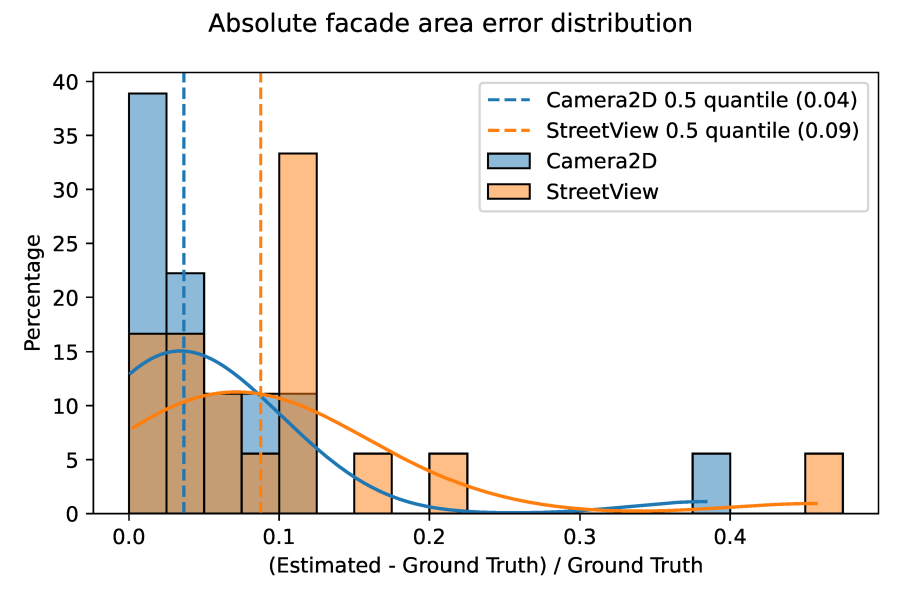}
\caption{\label{fig:Absolute_facade_area_error_distribution} Distribution of absolute errors in facade area estimations. }
\end{figure}

\paragraph{Areas of the windows and facades}
We evaluate the estimated area of the windows and facade by the normalized absolute error as shown in Figure~\ref{fig:Absolute_window_area_error_distribution} and Figure~ \ref{fig:Absolute_facade_area_error_distribution}, respectively. The median (50\% quantile) of the error is 9\% for StreetView and 4\% for Camera2D.
It is expressed as a ratio of the absolute error to the actual value.
As we can see that the performances for window area estimation are similar between Camera2D and StreetView, whereas Camera2D outperforms StreetView for facade area estimation due to the targeted data collection process, which reduces issues such as occlusion.

\begin{figure}[ht!]
\centering
\includegraphics[width=0.8\linewidth]{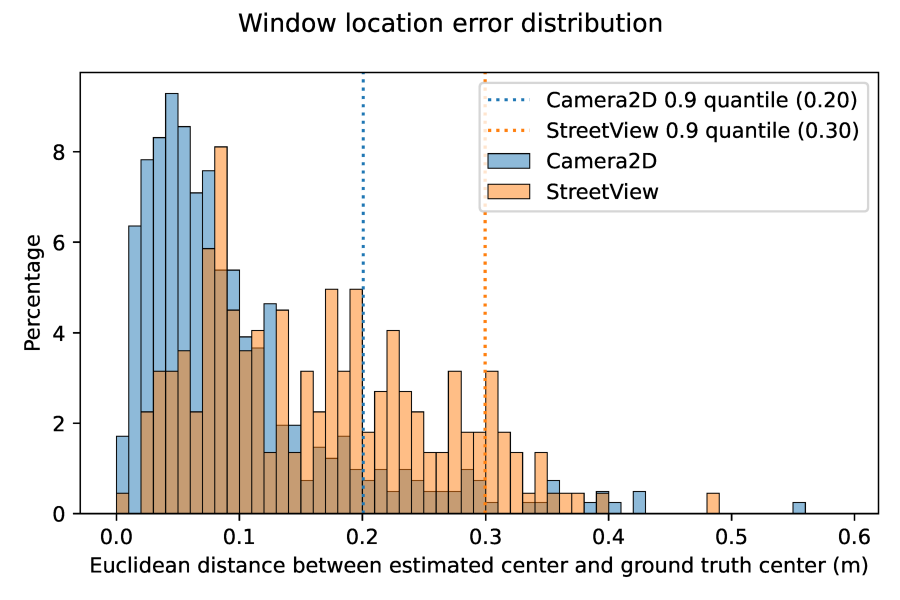}
\caption{\label{fig:Window_location_error_distribution} Distribution of positional errors for window detections across different building facades. }
\end{figure}

\begin{figure}[ht!]
\centering
\includegraphics[width=0.8\linewidth]{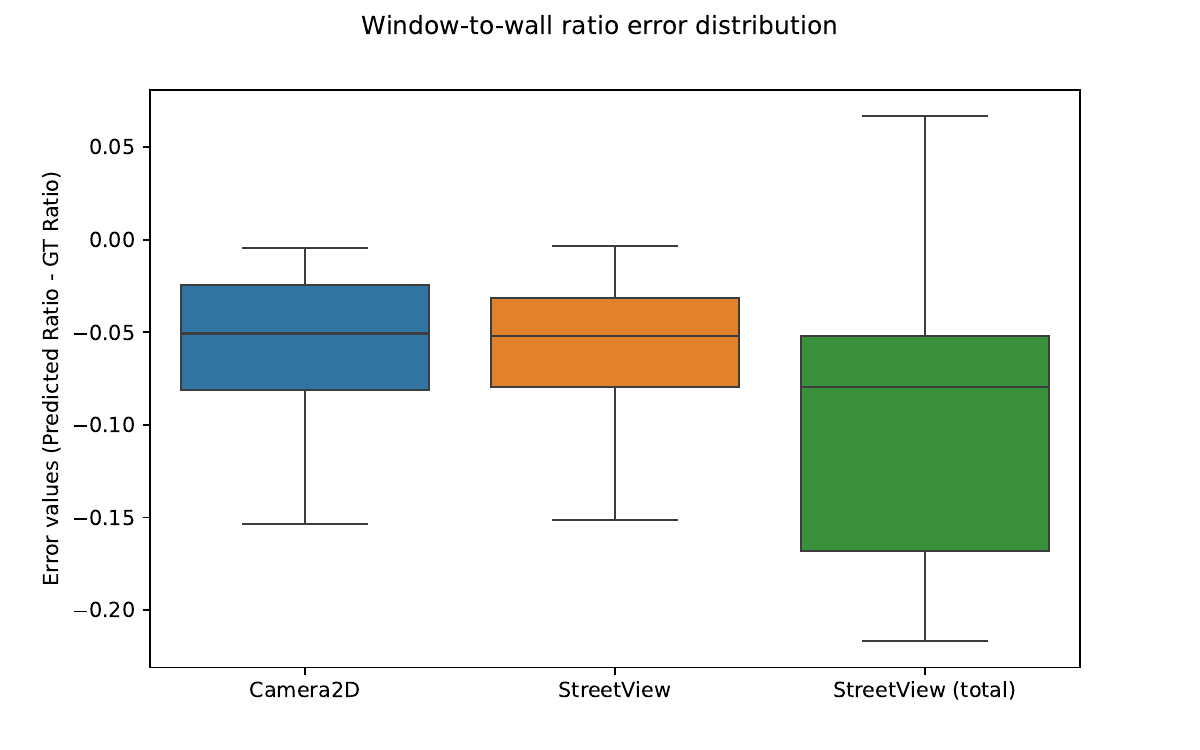}
\caption{\label{fig:Window-to-wall_ratio_error_distribution} Box plot of the aggregated window-to-wall ratio error (cf.~\ref{sec:experimental_setup}) across various buildings. }
\end{figure}

\paragraph{Location of the windows}

As we can see from Figure~\ref{fig:Window_location_error_distribution}, compared to the StreetView pipeline, Camera2D offers an advantage in terms of spatial accuracy due to its use of a fully 3D model for buildings. By constructing and utilizing 3D models, this pipeline more effectively captures the 3D structure of building facades. This results in a more faithful representation of the physical world and hence minimizes parallax errors.

The StreetView pipeline relies on the assumption that all planes are flat, which can lead to significant parallax issues when these planes are viewed from oblique angles. The larger the angle of observation relative to the plane, the more severe the parallax error. This limitation is particularly visible in urban settings where the angle of data capture can vary widely.

The area and position errors, though higher for StreetView, remain moderate overall, particularly for buildings without extreme occlusion or long facades. Most of the inaccuracies come from slight shifts or scaling issues of detected windows rather than gross misdetections. In other words, the methods are generally able to identify the approximate window locations and sizes, but fine-grained geometric precision suffers under sparse or occluded conditions.
\paragraph{WWR}
As shown in Figure~\ref{fig:Window-to-wall_ratio_error_distribution} and Table~\ref{tab:window_detection}, we present the error of the estimated WWR.
Camera2D and StreetView both yield comparable results in estimating the WWR with Camera2D performs slightly better.

The difference between StreetView and StreetView (Total) results highlights the critical impact of data completeness. Missing facades penalize StreetView Total F1-Scores and WWR errors significantly. This is because missing facade data is treated as having zero detected windows, dragging down global performance measures. However, when missing facades are accounted for through simple imputation (averaging WWR from detected facades), the WWR estimation improves substantially, illustrating that lightweight corrective strategies can partially compensate for sparse data gaps.

\begin{table}[htbp]
\centering
\caption{Benchmark comparison of 3D modeling approaches.}
\label{tab:benchmark}
\small
\resizebox{\textwidth}{!}{
\begin{tabular}{l c c c c}
\toprule
 Method & WWR Accuracy & \multicolumn{2}{c}{Time (per building)} & Scalability \\
 \cmidrule(lr){3-4}
  & (Typ. Error) & Collection & Processing & (Overall) \\
\midrule
 Conventional LiDAR & \textasciitilde 1 \% & \textasciitilde 120 min & \textasciitilde 6.0 hrs & Low \\
 SI3FP StreetView & \textasciitilde 6 \% & \textasciitilde 2 min & \textasciitilde 0.1 hrs & Very High \\
 SI3FP Camera2D & \textasciitilde 5 \% & \textasciitilde 30 min & \textasciitilde 2.5 hrs & High \\
\bottomrule
\end{tabular}
}
\end{table}

\paragraph{Scalability}

To address the scalability of the data collection, we summarize the comparison of different data sources (LiDAR, StreetView, and Camera2D) in Table \ref{tab:benchmark}.
LiDAR provides a dense point cloud and precise building coordinates through manual annotation but the data collection requires approximately 120 minutes and is costly. In contrast, StreetView, which captures panoramic images along with camera pose and plane definitions, offers a sparse data density but is significantly quicker and cost-effective, requiring around 2 minutes using a data collection vehicle. When such a data collection vehicle is not available, online services and resources can be utilized to achieve the data collection, with the cost being vendor-dependent.
Camera2D captures dense perspective camera images with local coordinate reference points and takes about 30 minutes with a relatively low cost with a simple camera setup. This comparison highlights the trade-offs between data density, equipment complexity, time efficiency, and cost for scalable thermal modeling.

\subsection{Computational Performance and Scalability Considerations}
Beyond accuracy, the practical applicability of the SI3FP pipeline depends on its computational requirements. Using the hardware detailed in Section~\ref{sec:experimental_setup} (NVIDIA RTX 3090 GPU, AMD Ryzen 7 5800X3D CPU, 64GB RAM), we provide indicative performance characteristics for key stages. A summary, including the computational complexity and theoretical scalability, is provided in Table~\ref{tab:computation_summary}. For convenience, we denote the number of panorama images as $N_{pano}$, facades as $N_f$, orthographic images as $N_{o}$, windows as $N_w$, planes as $N_{pl}$, and image resolution as $H \times W$. The number of detected lines is denoted by $N_{line}$ and detected boxes is $N_{box}$. For the COLMAP algorithm in C.2, we use $K$ to denote the number of adjacent images for each image in the sequential matching, and we use $N_{img}$ to denote the total images captured. Moreover, we use $N_{\text{pano}/f}$ to denote the number of panorama images per facade. Similarly, $N_{\text{pl}/\text{pano}}$ denotes the number of planes per panorama. In general, we use $N_{x/y}$ to denote the number of $x$ per $y$. We assume the detection time of ResNet-50 is constant per pixel. 

The Camera2D path involves the most computationally intensive steps per building. SfM (Step C.2 using COLMAP) typically required 2 hours, demanding significant RAM and benefiting from GPU acceleration for feature matching. NeRF modeling (Step C.3 using Instant-NGP) required approximately 10 minutes of training time on the GPU, with its VRAM usage scaling with scene complexity; subsequent orthographic rendering was fast (seconds per view). 

The StreetView path generally has a lower computational cost per building after the initial data acquisition (S.1). Plane clustering (S.2) and geometry-based facade detection (S.4) were relatively fast (seconds to minutes, CPU-bound). The primary computational step is often the multi-view image alignment (S.3), which took seconds to minutes per facade cluster on our hardware, its duration scaling with the number of views (N) to align and benefiting from GPU acceleration for parts like feature matching.
For both paths, the final deep learning inference for window detection (M.1) was efficient when using the GPU, processing each facade image in approximately 20-100 milliseconds.

Given these performance profiles, large-scale deployment for analyzing hundreds or thousands of buildings hinges on parallelization. The pipeline is well-suited for building-level parallelization, where each building is processed independently on separate compute resources. Comparing the two paths, the StreetView approach, despite potential bottlenecks in alignment for facades seen from very many views, generally offers higher throughput for large building stocks due to lower per-building compute demands (assuming data availability). The Camera2D path, while providing detailed NeRF models, requires significantly more GPU time and memory per building, making it better suited for targeted analysis or requiring substantial parallel computing infrastructure if applied at a very large scale.

\begin{table}[htbp!]
\centering
\caption{Estimated computational requirements and scalability for SI3FP pipeline steps. }
\label{tab:computation_summary} 
\resizebox{\textwidth}{!}{
\begin{tabularx}{\textwidth}{l L l L} 
\toprule
 Step & Resource(s) & \shortstack{Time \\ (per building)} & Scalability (per building)\\
\midrule
 \multicolumn{4}{l}{\textit{StreetView Path Steps}} \\
 S.1 & Network, CPU (Low) & \textasciitilde 2 min & $\mathcal{O}(N_{pano})$ \\
 S.2 & CPU (Multi-core), RAM & $<$ 1 min & $\mathcal{O}(N_{pano}^2N_{pl/pano}^2)$\\
 S.3 & CPU, GPU & $<$ 1 min & $\mathcal{O}(N_fN_{pano/f}^2WH)$ \\
 S.4 & CPU & $<$ 10 s & $\mathcal{O}(N_{line}^2)$\\
 \midrule
 \multicolumn{4}{l}{\textit{Camera2D Path Steps}} \\
 C.1 & Manual Effort & \textasciitilde 30 min & $\mathcal{O}(N_{img})$ \\
 C.2 & CPU, GPU, High RAM & \textasciitilde 2 h &  \textasciitilde$O(N_{img}^3)$ (worst case), depends on $K$ and $HW$ \\
 C.3 & GPU (High VRAM), RAM & $<$ 10 min & Training complexity of NeRF\\
 \midrule
 \multicolumn{4}{l}{\textit{Merged Steps}} \\
 M.1 & GPU (VRAM) & \textasciitilde 1 s & $\mathcal{O}(N_oHW+N_{box}^2)$\\
 M.2 & CPU (Low) &\textasciitilde  1 s & $\mathcal{O}(N_f+N_w)$  \\
\bottomrule
\end{tabularx}
}
\end{table}

\section{Discussion}

This paper explored different methods for generating building models suitable for early-stage energy simulation. Our proposed pipeline, along with alternative approaches, proved effective at capturing the thermal envelope -- including roofs, floors, and walls -- with sufficient accuracy for practical applications. 

\begin{figure}[h!]
\centering
\includegraphics[width=1\linewidth]{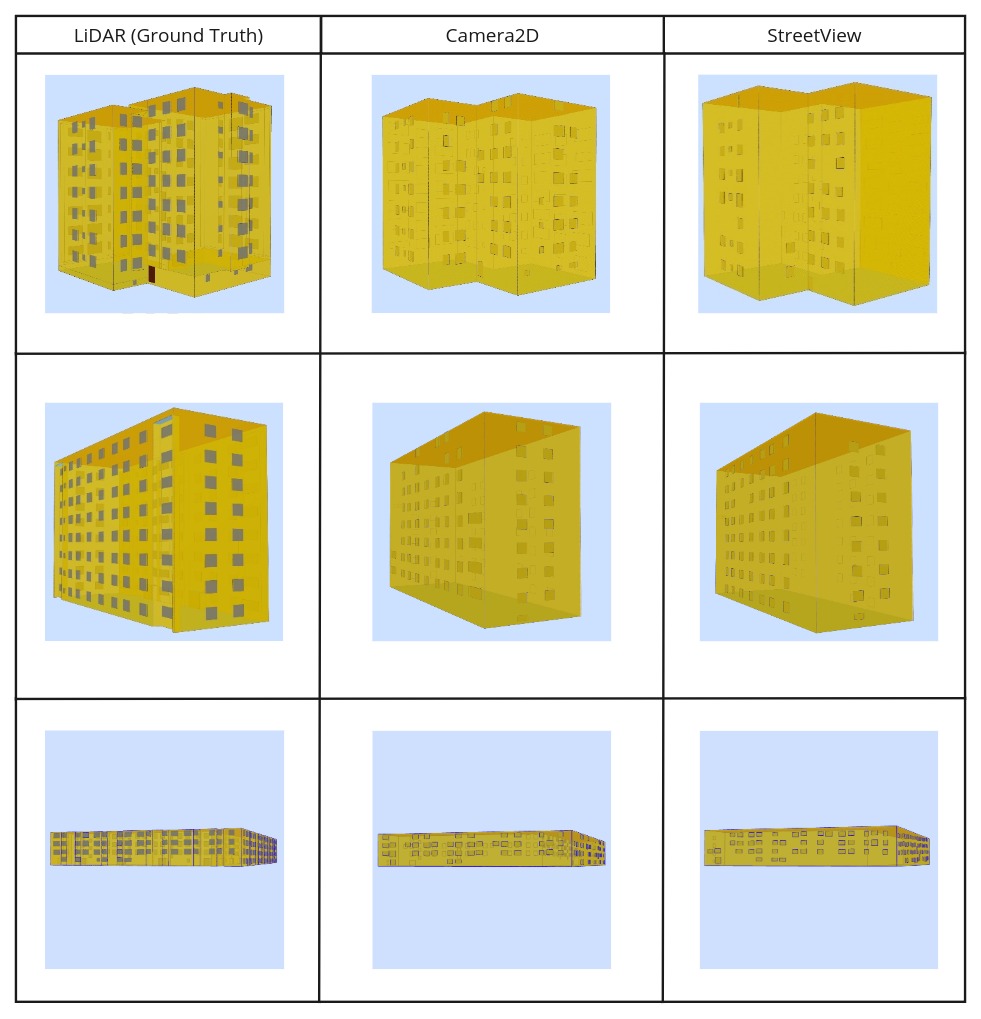}
\caption{\label{fig:example}  Visual comparison between the different models obtained by each method, 3D LiDAR scanned (used as ground truth), StreetView, and Camera2D.}
\end{figure}

However, challenges remain in accurately detecting windows using data collected from budget sensors at scale, reflecting the core difficulty of balancing scalability and accuracy. The visual comparison in Figure~\ref{fig:example} highlights that no data source is flawless. For example, StreetView highly depends on the availability of unobstructed views towards the building; obstructions will decrease the amount of data that can be obtained. With dense image capturing methods, data collection from ground level severely limits the amount of data for the top floors.
Additionally, the presence of balconies creates obstructions, making it difficult to accurately identify the dimensions or full shape of windows.

\paragraph{Limitation and future work}
We acknowledge several limitations in our current paper.
First, regarding geometric fidelity and representation, projecting facades onto a single plane, as done in both the sparse and dense paths, can distort off-plane geometries such as curves and deep ornamentation. This design directly supports the objectives of our paper: for early-stage renovation analysis, the focus is on large-scale applicability rather than fine-grained modeling. Capturing complex geometries would require pre-segmentation into multiple planes, significantly increasing computational cost, annotation burden, and system complexity. Similarly, the current bounding-box-based window detection (Step M.1) may not capture non-rectangular windows precisely, but it provides a scalable and effective solution for standard facade layouts. For applications requiring high-fidelity architectural reconstruction or conservation work, future extensions could integrate instance segmentation and multi-plane modeling.
Second, challenging scene properties such as glass facades and occlusions remain difficult. Reflective and transparent materials degrade camera-based reconstruction, and dense urban clutter limits complete facade visibility. Although the StreetView ensemble strategy (Algorithm~\ref{alg:detection_fusion}) partially mitigates these effects, fully addressing them would require multi-modal sensing or specialized modeling, which are promising directions for future research.
Third, data acquisition constraints introduce limitations. Ground-level captures yield oblique views of upper facades, reducing resolution and orthographic clarity at height. Incorporating aerial imagery, such as UAV-based oblique views, would enhance coverage for tall buildings. Furthermore, while the incremental SfM stage (Step C.2) provides reliable reconstruction, it can be slow and prone to drift on large datasets. Future improvements include exploring global SfM approaches, such as GLOMAP~\citep{10.1007/978-3-031-73661-2_4}, and integrating learned feature extractors and matchers to enhance scalability and robustness.
Additionally, there is currently a lack of public benchmark datasets tailored for facade parsing and thermal modeling, limiting standardized evaluation across studies. Developing such datasets would be an important direction to advance the field.

Overall, our system was designed to balance accuracy, scalability, and computational feasibility, targeting practical early-phase renovation needs. Although we tested our approach only on Swedish buildings, their geometry and typology are representative of building stocks common throughout Europe, making the method likely transferable to broader European contexts. Future work will extend the system toward high-fidelity reconstruction as data availability and computational resources improve.

It is worth noting that while many factors influence simulation accuracy (e.g., infiltration rates, material properties, internal loads), geometric accuracy -- particularly surface areas and volumes -- is especially critical. Errors in modeled area or volume inevitably lead to large distortions in energy simulation results. Encouragingly, for opaque envelope elements, our method and alternatives achieve high dimensional accuracy. For transparent surfaces, underestimation of window areas remains an issue. In addition, missing facade from sparse data collection limits the detection capacity. This could be addressed in future work through pattern analysis techniques, although such synthesis lies beyond the current scope, which focuses on capturing existing building data rather than creating inferred models.

\section{Conclusion}
This paper presented SI3FP, a modular, scalable pipeline for generating thermal 3D models of buildings from visual data, designed to support early-stage renovation analysis. By integrating two complementary paths -- StreetView for scalable, sparse data collection and Camera2D for targeted, high-resolution modeling -- our system offers users flexibility in balancing cost, effort, and accuracy. In both paths, we apply orthographic transformation to create a consistent image domain for facade feature detection and geometry parameterization. This unified semantic interface reduces perspective distortion, simplifies geometry extraction, and enables modular reuse of downstream methods. 

The technical contributions of SI3FP include a direct orthographic image generation approach that avoids projection artifacts, a multi-view geometric plane clustering and alignment method, and a tailored RANSAC-based facade extraction technique. These are combined with a unified interface for semantic parsing and 3D thermal modeling, resulting in an end-to-end system that maintains an approximate error of ~5\% in WWR estimation -- well within the acceptable range for early-phase renovation planning.

Beyond its practical utility, SI3FP contributes several conceptual insights: \\
1) \textbf{Multi-view fusion improves sparse data robustness:} Rather than selecting a single optimal image per facade, SI3FP’s ensemble-based multi-view strategy improves resilience against occlusion and poor lighting, especially important in real-world street-level imagery. \\
2) \textbf{Rendering is not sufficient for 3D modeling -- parameterization is a key design choice:} Advanced image-to-3D rendering methods such as NeRF represent an important step toward automation. They enable data-rich, photorealistic reconstructions from unstructured images. However, rendering alone does not yield structured 3D geometry. To transition from rendering to actionable 3D models, \emph{parameterization} is required, i.e., representing geometry as polylines, planes, or meshes. This is a design choice: the more granular the parameterization (e.g., arbitrarily shaped polygons, dense meshes), the more complex geometry it can represent -- but also the more complexity and lower scalability it brings. Coarser primitives like rectangles or planes are easier to manage and often sufficient in early-stage modeling. The appropriate parameterization should be selected based on the use case, e.g., detailed facade conservation may justify mesh-based modeling, while large-scale energy planning may not. This is a design choice that needs to be carefully made jointly by the technical developers and end users.
3) \textbf{Task-specific modeling precision can be optimized without full 3D mesh reconstruction:} Our results show that accurate window detection and WWR estimation do not require full 3D mesh models. Instead, structured abstractions derived from image-plane primitives (e.g., bounding boxes, planes) can offer sufficient accuracy for thermal modeling at a much lower computational cost.

While SI3FP achieves satisfactory performance across diverse building geometries, several limitations remain that offer promising avenues for future work:
1) \textbf{Generalizing to more complex geometries:} Current plane-based simplification is insufficient for heavily curved facades or articulated surfaces. While not within the scope of our primary focus, it may be beneficial for other applications. 
Integrating multi-plane, polygonal, or spline-based representations -- or even implicit surfaces such as signed distance functions -- could increase expressiveness. However, as noted above, this increased expressiveness comes with added complexity. Therefore, exploring scalable algorithmic solutions is a promising direction for such use cases.
2) \textbf{From rendering to modeling:} Building on the insight above, future work could explore hybrid pipelines that integrate NeRF or other neural rendering techniques with parametric fitting algorithms to automatically extract geometry. This includes polygon fitting, mesh extraction, or rule-based facade grammar parsing.
3) \textbf{Physics-informed modeling:} Incorporating thermal priors or physically based constraints into the detection and modeling stages could better inform the placement and sizing of windows, especially in edge cases.
4) \textbf{Citizen-contributed and crowdsourced data:} To support continental-scale building assessments, future systems could integrate community-contributed photos and drone footage, extending SI3FP with scalable data collection and automated labeling and validation.

Beyond renovation potential analysis, the generated 3D models open up opportunities for broader applications, including 3D modeling for daylight analysis and construction material mapping for urban mining, reuse, and broader circular economy initiatives.

\section{CRedit Author statement} 

Yinan Yu: Conceptualization, Methodology, Data curation, Software, Visualization, Validation, Writing - Original draft preparation, Writing - Reviewing and Editing;
Alex Gonzalez-Caceres: Conceptualization, Data curation, Validation, Writing - Original draft preparation;
Samuel Scheidegger: Data curation, Visualization, Software, Validation, Writing - Reviewing and Editing;
Sanjay Somanath: Conceptualization;
Alexander Hollberg: Conceptualization, Writing - Reviewing and Editing.

\section{Acknowledgements}
This work has been funded by Sweden’s Innovation Agency Vinnova through the project DecarbonAIte [grant number 2021-02759] and it is part of the Digital Twin Cities Centre supported by Sweden’s Innovation Agency Vinnova [grant number 2019-00041]. We would like to thank Dag Wästberg and Bernd Ketzler for their support in the project and Andreas Skälegård for providing data on the case study buildings. Furthermore, we would like to thank Jieming Yan for the support in the point cloud annotation. This work has also been funded by Chalmers Energy Area of Advance (AoA).

\bibliographystyle{elsarticle-num} 
\bibliography{references}

\end{document}